\documentclass{article}

  \usepackage[main, final]{neurips_2026}

\usepackage[utf8]{inputenc} 
\usepackage[T1]{fontenc}    
\usepackage{hyperref}       
\usepackage{url}            
\usepackage{booktabs}       
\usepackage{amsfonts}       
\usepackage{nicefrac}       
\usepackage{microtype}      
\usepackage{xcolor}         
\usepackage{xpatch}
\newtheorem{theorem}{Theorem}
\usepackage{amsmath}
\usepackage{bm}    
\usepackage{array}
\usepackage{stfloats}
\usepackage{balance}
\usepackage{float}
\usepackage{graphicx}
\usepackage{multirow}

\usepackage{algorithmic}
\usepackage{array}
\usepackage{stfloats}
\usepackage{balance}
\usepackage{float}

\usepackage{xpatch}

\title{C-LoRA: Continual Low-Rank Adaptation for Pre-trained Visual Models}

\author{%
	Xin Zhang \\
	Institute of Intelligent Information Processing \\
	Shanxi University \\
	Taiyuan 030006, China \\
	\texttt{zhangxin8572@sxu.edu.cn}
	\And
	Liang Bai\thanks{Corresponding author.} \\
	Institute of Intelligent Information Processing \\
	Shanxi University \\
	Taiyuan 030006, China \\
	\texttt{bailiang@sxu.edu.cn}
	\And
	Xian Yang \\
	Alliance Manchester Business School \\
	The University of Manchester \\
	Manchester M13 9PL, United Kingdom \\
	\texttt{xian.yang@manchester.ac.uk}
}

\begin{document}

\maketitle

\begin{abstract}
  Pre-trained visual models have become fundamental in computer vision, but they face challenges in continual learning scenarios where data and tasks evolve over time. Low-Rank Adaptation (LoRA) offers efficient fine-tuning capabilities but remains limited for such dynamic environments. 
  Standard LoRA cannot distinguish important subspaces, causing critical knowledge to be overwritten in sequential training. 
  Existing approaches address this by dynamically expanding the set of LoRA adapters—either maintaining a growing pool of task-specific modules or merging new adapters into prior ones—at the cost of unbounded parameter growth or increasing inference complexity. 
  We propose Continual Low-Rank Adaptation (C-LoRA), a method that enables a single, shared LoRA adapter to handle sequential tasks without catastrophic forgetting—without requiring any module selection or fusion at inference. The core of C-LoRA is a learnable routing matrix $\boldsymbol{\mathcal{R}}$ that explicitly controls how each rank-one subspace contributes to the weight update. This matrix is decomposed into a stability component ($\boldsymbol{\mathcal{R}}_{\text{base}}$), which preserves knowledge from prior tasks, and a plasticity component ($\boldsymbol{\mathcal{R}}_{\delta}$), which drives adaptation to the current task—providing direct control over the stability-plasticity trade-off. We analyze how $\boldsymbol{\mathcal{R}}$ governs gradient flow during sequential training, and demonstrate competitive performance across multiple benchmarks. The code is available at \url{https://github.com/lambor9973/C-LoRA}.
  
\end{abstract}

\section{Introduction}
Pre-trained visual models~\cite{dosovitskiy2020image, wang2022dualprompt,wang2022learning} have revolutionized computer vision applications, but they face fundamental limitations when deployed in dynamic environments where tasks and data distributions continuously evolve~\cite{parisi2019continual}. This challenge, known as continual learning~\cite{wang2024comprehensive}, requires models to progressively accumulate knowledge while preserving performance on previously learned tasks.

Low-Rank Adaptation (LoRA)~\cite{hu2021lora} is a parameter-efficient fine-tuning method that adapts large-scale pre-trained models to downstream tasks by introducing low-rank matrices to update a small subset of parameters while keeping the core pre-trained weights frozen. This approach preserves the general knowledge captured during pre-training, reduces computational and storage costs, and mitigates the risk of overfitting by limiting updates to task-specific components~\cite{ayoub2023requirements}. LoRA has proven highly effective in fields such as computer vision, where pre-trained models are commonly deployed for diverse and evolving applications~\cite{yu2024boosting,zhou2024expandable,aljundi2017expert,hu2023dense}.

An overlooked question is why applying single LoRA to a task sequence leads to rapid forgetting, even when the adapter has sufficient capacity. We identify the cause: the standard update $\Delta\mathbf{W} = \mathbf{AB}$ implicitly assigns a uniform weight to every rank-one subspace $\mathbf{a}_i\mathbf{b}_i^\top$, providing no mechanism to distinguish subspaces that encode critical past knowledge from those that are underutilized. As a result, all subspaces are equally exposed to gradient updates from new tasks, regardless of their importance. 
Introducing a learnable routing matrix $\boldsymbol{\mathcal{R}} \in \mathbb{R}^{r \times r}$ into the update, giving $\Delta\mathbf{W} = \mathbf{A}\boldsymbol{\mathcal{R}}\mathbf{B}$, addresses this by allowing the model to learn differential subspace weights. However, Theorem~\ref{thm:1} shows that the gradients flowing into the shared bases $\mathbf{A}$ and $\mathbf{B}$ scale with $\|\boldsymbol{\mathcal{R}}\|_2$. As knowledge accumulates, $\|\boldsymbol{\mathcal{R}}\|_2$ grows, and the subspaces with the largest routing weights—those most critical for past tasks—experience the strongest gradient perturbations during new-task training. The subspaces most critical for stability are paradoxically the most vulnerable to disruption. 

Prior approaches treats this conflict as a capacity problem: methods such as~\cite{zhou2024expandable, liang2024inflora, wu2025sd} expand the adapter pool with each new task—either retaining task-specific modules separately or merging them incrementally—at the cost of linear parameter growth and additional inference overhead from module selection or fusion. The implicit assumption is that stability and plasticity cannot be jointly satisfied within a fixed low-rank space. We show the conflict stems not from low-rank capacity, but from coupled gradient amplification that disproportionately disrupts critical subspaces. Resolving that mechanism directly makes expansion unnecessary.

We propose Continual Low-Rank Adaptation (C-LoRA), which mitigates the stability-plasticity conflict within a single, fixed-size adapter. C-LoRA decomposes $\boldsymbol{\mathcal{R}} = \boldsymbol{\mathcal{R}}_{\text{base}} + \boldsymbol{\mathcal{R}}_{\delta}$: the stability component $\boldsymbol{\mathcal{R}}_{\text{base}}$ accumulates knowledge from prior tasks and is excluded from gradient flow during new-task training; the plasticity component $\boldsymbol{\mathcal{R}}_{\delta}$ drives adaptation to the current task without disrupting consolidated knowledge. This decoupling breaks the conflict identified with theoretical guarantees and validated empirically across multiple benchmarks.

\section{Related Work}

Continual learning requires models to accumulate knowledge across a sequence of tasks without forgetting~\cite{wang2024comprehensive}. Existing methods fall into four broad categories. Replay-based methods~\cite{liu2021rmm,luo2023class,wu2019large,buzzega2020dark,wang2022memory,wang2021triple,rebuffi2017icarl} store a subset of past data and interleave it with new-task training. Knowledge distillation methods~\cite{dhar2019learning,douillard2020podnet,li2017learning,simon2021learning,tao2020topology} preserve the outputs of prior models via auxiliary losses. Architecture-based methods~\cite{shi2022mimicking,zhao2020maintaining,hu2023dense,wang2023incorporating,yang2022continual} allocate task-specific parameters within an expanding network. Regularization-based methods~\cite{zhou2023hierarchical,aljundi2018memory,zenke2017continual,dhar2019learning,li2017learning} penalize changes to weights deemed important for past tasks.

LoRA~\cite{hu2021lora} achieves parameter-efficient adaptation by inserting trainable low-rank matrices into frozen pre-trained weights. It performs well in static settings~\cite{hayou2024lora+,li2023loftq,kopiczko2023vera,fan2025make}, but applying a single LoRA adapter to a task sequence leads to rapid forgetting. The dominant response has been to expand the adapter set: InfLoRA~\cite{liang2024inflora} and EASE~\cite{zhou2024expandable} maintain a growing pool of task-specific modules; SD-LoRA~\cite{wu2025sd} introduces a new module per task and merges it into the previous one. These strategies treat the underlying conflict as a capacity problem—implicitly assuming that stability and plasticity cannot be jointly satisfied within a fixed low-rank space. Prior work has shown more broadly that parameter isolation strategies limit backward knowledge transfer by treating new-task updates as interference rather than as a source of refinement~\cite{lin2022beyond}. The resulting systems also incur linear parameter growth and additional inference overhead from module selection or fusion.

Rather than expanding parameters, C-LoRA identifies and resolves it through a decomposed routing matrix within a single, fixed-size adapter. This shifts the problem from parameter allocation to gradient flow control.

\section{Methodology}

\subsection{Preliminaries}
For a pre-trained linear layer $\mathbf{W_0} \in \mathbb{R}^{d \times k}$, LoRA applies a low-rank update $\Delta \mathbf{W} = \mathbf{AB}$, where $\mathbf{A} \in \mathbb{R}^{d \times r}$ and $\mathbf{B} \in \mathbb{R}^{r \times k}$ are trainable matrices with rank $r \ll \min(d, k)$. The updated weight for task $t$ is:
\begin{equation}
	\mathbf{W}_t = \mathbf{W}_0 + \Delta \mathbf{W}_t = \mathbf{W}_0 + \mathbf{A}_t \mathbf{B}_t,
\end{equation}
where $\mathbf{A}_t$ and $\mathbf{B}_t$ are the task-specific low-rank matrices. By keeping $\mathbf{W}_0$ frozen and only updating $\mathbf{A}_t$ and $\mathbf{B}_t$, LoRA achieves significant parameter efficiency compared to full fine-tuning.

In continual learning, existing LoRA-based methods~\cite{wei2024online,zhou2024expandable,liang2024inflora} introduce a separate update $\Delta\mathbf{W}_t = \mathbf{A}_t\mathbf{B}_t$ per task, maintaining a growing collection:
\begin{equation}
	\{(\mathbf{A}_1, \mathbf{B}_1), (\mathbf{A}_2, \mathbf{B}_2), \dots, (\mathbf{A}_T, \mathbf{B}_T)\}.
\end{equation}
However, this approach treat the 
stability-plasticity conflict as a capacity problem rather than addressing its cause. We 
instead identify the cause within the structure of the update itself.

\subsection{Generalized Low-Rank Adaptation via a Routing Matrix}

\begin{figure*}[!t]
	\centering
	\begin{minipage}{\textwidth}
		\centering
		\includegraphics[width=\textwidth, trim=2cm 22.5cm 12cm 0cm, clip]{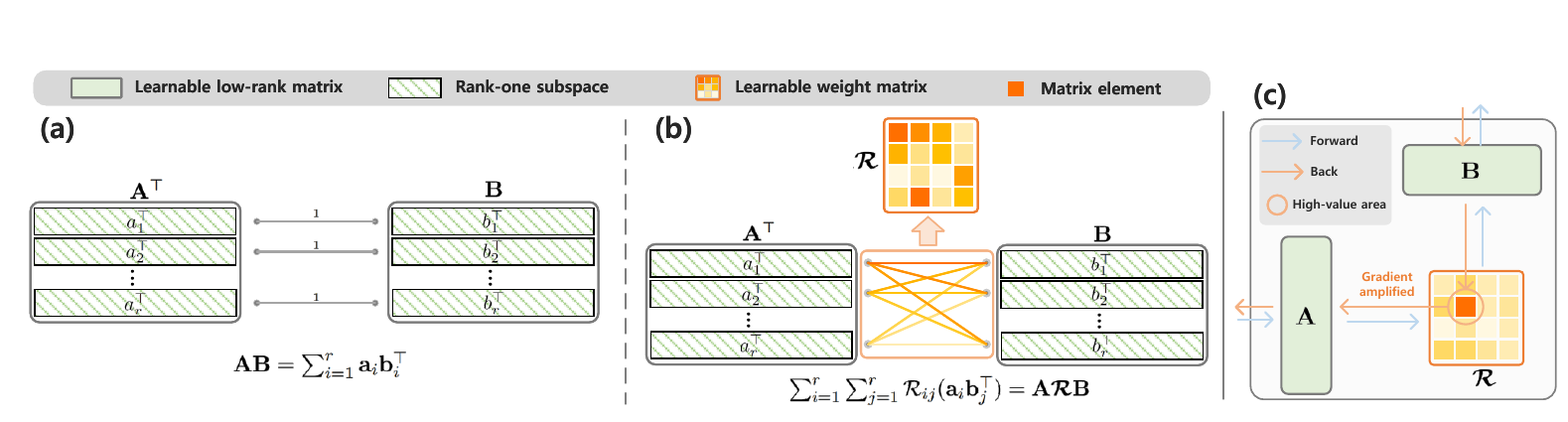}
	\end{minipage}%
	\caption{(a) The standard LoRA formulation implicitly assumes fixed, uniform weights (implicitly 1) and only allows interaction between corresponding rank-one subspaces ($a_i b_i^\top$). (b) C-LoRA introduces a learnable routing matrix $\boldsymbol{\mathcal{R}}$, enabling fine-grained control over subspace contributions and capturing complex cross-subspace interactions between $a_i$ and $b_j$. (c) The inherent conflict in continual learning: High-magnitude areas in $\boldsymbol{\mathcal{R}}$ (hotspots) are crucial for stability (preserving past knowledge), but they simultaneously amplify the gradients flowing to the shared matrices $\mathbf{A}$ and $\mathbf{B}$ during backpropagation, thereby risking catastrophic forgetting.}
	\label{fig:1} 
\end{figure*}

The update $\Delta\mathbf{W} = \mathbf{AB}$ decomposes as a sum of $r$ rank-one outer products:
\begin{equation}
	\Delta\mathbf{W} = \sum_{i=1}^{r} \mathbf{a}_i \mathbf{b}_i^\top,
\end{equation}
where $\mathbf{a}_i$ is the $i$-th column of $\mathbf{A}$ and $\mathbf{b}_i^\top$ is the $i$-th row of $\mathbf{B}$. As shown in Figure~\ref{fig:1}(a), each subspace 
$\mathbf{a}_i\mathbf{b}_i^\top$ contributes with a uniform, fixed weight of one. This leaves the adapter unable to differentially protect subspaces as knowledge accumulates.

Relaxing this constraint by introducing a learnable coefficient, $\mathcal{R}_{ij}$, for each potential interaction between a basis vector $\mathbf{a}_i$ and $\mathbf{b}_j$, as shown in Figure~\ref{fig:1}(b). This leads to a generalized formulation: 
\begin{equation}
	\Delta \mathbf{W} = \sum_{i=1}^{r} \sum_{j=1}^{r} \mathcal{R}_{ij} (\mathbf{a}_i \mathbf{b}_j^\top)= \mathbf{A}\boldsymbol{\mathcal{R}}\mathbf{B},
\end{equation}
where $\mathcal{R}_{ij}$ independently controls the contribution of each cross-subspace component $\mathbf{a}_i\mathbf{b}_j^\top$ (Figure~\ref{fig:1}(b)). Figure~\ref{fig:2} confirms that $\boldsymbol{\mathcal{R}}$ develops structured, non-uniform weights during training: initially near-uniform, the matrix develops localized high-magnitude regions and increasing element variance, indicating learned subspace prioritization.

\subsection{Resolving the Stability-Plasticity Conflict via Decoupled Adaptation}


Let \(\theta_t\) represent the model parameters after learning task \(t\). The metric for stability \(S(\theta_t, \theta_{t+1})\) quantifies the degradation in performance on previous tasks due to the new adaptation:
\begin{equation}
	S(\theta_t, \theta_{t+1}) = \mathbb{E}_{D_{old}}[\mathcal{L}(\theta_{t+1}) - \mathcal{L}(\theta_t)].
\end{equation}
Minimizing \(S\) is crucial for mitigating catastrophic forgetting. Our analysis relies on standard assumptions in optimization theory: the loss function \(\mathcal{L}\) is \(L\)-smooth and the parameters are bounded (e.g., \(\|\mathbf{A}\|_F, \|\mathbf{B}\|_F \leq M\)).

\textbf{The amplification problem.} The gradients for the shared bases under 
$\mathbf{A}\boldsymbol{\mathcal{R}}\mathbf{B}$ are:


\begin{equation}
	\begin{aligned}
		\frac{\partial \mathcal{L}}{\partial \mathbf{A}} = \boldsymbol{G} (\boldsymbol{\mathcal{R}}\mathbf{B})^\top = \boldsymbol{G} \mathbf{B}^\top \boldsymbol{\mathcal{R}}^\top, \\
		\frac{\partial \mathcal{L}}{\partial \mathbf{B}} = (\mathbf{A}\boldsymbol{\mathcal{R}})^\top \boldsymbol{G} = \boldsymbol{\mathcal{R}}^\top \mathbf{A}^\top \boldsymbol{G},
	\end{aligned}
\end{equation}

where $\boldsymbol{G} = \frac{\partial \mathcal{L}}{\partial \mathbf{W}}$ is the upstream gradient. A large $|\mathcal{R}_{ij}|$ therefore amplifies the gradients flowing into $\mathbf{A}$ and $\mathbf{B}$, as formalized below.


\begin{figure*}[!t]
	\centering
	\begin{minipage}{0.9\textwidth}
		\centering
		\includegraphics[width=\textwidth, trim=2.5cm 10cm 2cm 2.5cm, clip]{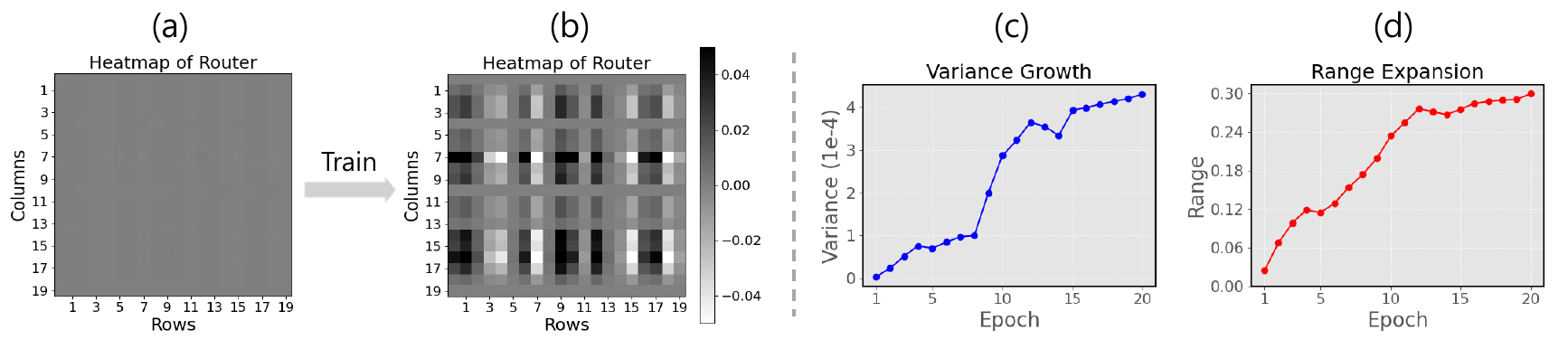}
	\end{minipage}%
	\caption{\textbf{(a):} Heatmap of the $\boldsymbol{\mathcal{R}}$ matrix at initialization, showing a nearly uniform distribution;\textbf{(b):} Heatmap of the $\boldsymbol{\mathcal{R}}$ matrix after training, with noticeable localised value changes; \textbf{(c):} Variance of matrix elements increasing with training epochs; \textbf{(d):} Range of matrix values expanding as training progresses.}
	\label{fig:2} 
\end{figure*}

\begin{theorem} \label{thm:1} Assume the norms of the shared matrices and the upstream gradient are bounded, i.e., \(\|\mathbf{A}\|_F, \|\mathbf{B}\|_F \leq M\) and \(\|\textbf{G}\|_F \leq G_{\text{max}}\). In a joint update scenario, the magnitude of the gradients flowing to the shared matrices \(\mathbf{A}\) and \(\mathbf{B}\) is directly bounded by the norm of the routing matrix \(\boldsymbol{\mathcal{R}}\). Specifically, \begin{equation} \left\| \frac{\partial \mathcal{L}}{\partial \mathbf{A}} \right\|_F \leq G_{\text{max}} M \|\boldsymbol{\mathcal{R}}\|_2, \quad \left\| \frac{\partial \mathcal{L}}{\partial \mathbf{B}} \right\|_F \leq G_{\text{max}} M \|\boldsymbol{\mathcal{R}}\|_2. \end{equation} \end{theorem}

\begin{figure*}[!t]
	\centering
	\begin{minipage}{0.95\textwidth}
		\centering
		\includegraphics[width=\textwidth, trim=1cm 8cm 2cm 2cm, clip]{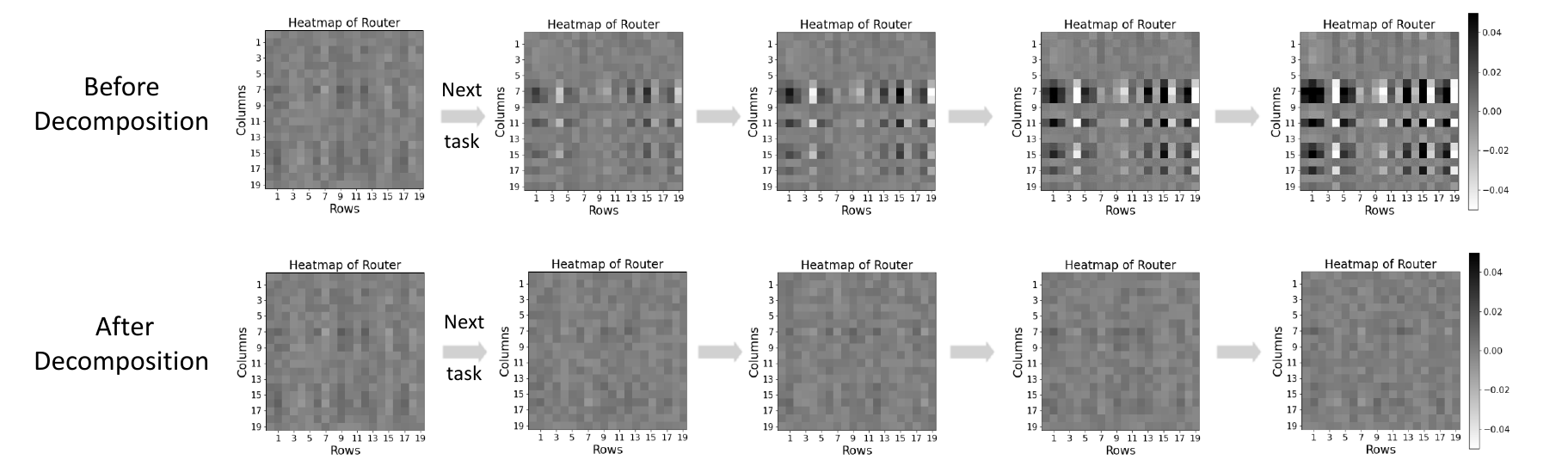}
	\end{minipage}%
	\caption{Evolution of the routing matrix $\boldsymbol{\mathcal{R}}$ heatmap during 5-task class-incremental learning. \textbf{Top Row (Before Decomposition):} Without decomposition, $\boldsymbol{\mathcal{R}}$ values progressively intensify in specific local regions. This highlights that subspaces critical for past tasks continue to receive strong updates when learning new tasks. \textbf{Bottom Row (After Decomposition):} With decomposition, updates primarily occur in the task-specific $\boldsymbol{\mathcal{R}}_{\delta }$, resulting in changes to $\boldsymbol{\mathcal{R}}$ that are more distributed across the matrix rather than being highly localized.
	}
	\label{fig:3} 
\end{figure*}

As tasks accumulate, $\|\boldsymbol{\mathcal{R}}\|_2$ 
grows, directly increasing the gradient magnitude on $\mathbf{A}$ and $\mathbf{B}$. 
The subspaces most important for stability—those with the largest 
$|\mathcal{R}_{ij}|$—are simultaneously the most susceptible to disruption. 
Figure~\ref{fig:3} (top row) visualizes this: without intervention, high-magnitude 
regions in $\boldsymbol{\mathcal{R}}$ act as hotspots for destructive updates in 
subsequent tasks.

\textbf{Decoupled routing.} The amplification arises because $\boldsymbol{\mathcal{R}}$ simultaneously carries consolidated knowledge and modulates gradient flow. We decouple these roles by decomposing:
\begin{equation}
	\mathbf{\boldsymbol{\mathcal{R}} = \boldsymbol{\mathcal{R}}_{\text{base}} + \boldsymbol{\mathcal{R}}_{\delta}}
\end{equation}

$\boldsymbol{\mathcal{R}}_{\text{base}}$ accumulates knowledge from all prior tasks and is treated as a constant during new-task training—gradient flow through it is stopped. $\boldsymbol{\mathcal{R}}_{\delta}$ is the trainable component for the current task. The update becomes:

\begin{equation}
	\Delta\mathbf{W}' = \phi(\mathbf{A \boldsymbol{\mathcal{R}}_{\text{base}} B}) + \mathbf{A \boldsymbol{\mathcal{R}}_{\delta } B},
\end{equation}
where $\phi(\cdot)$ denotes the stop-gradient operation. Gradients into $\mathbf{A}$  and $\mathbf{B}$ are now modulated by $\boldsymbol{\mathcal{R}}_{\delta}$ alone, whose norm is small at the start of each task. This directly breaks the amplification mechanism in Theorem~\ref{thm:1}. Figure~\ref{fig:3} (bottom row) confirms that updates become distributed rather than concentrated at hotspots. After training,  knowledge is consolidated: $\boldsymbol{\mathcal{R}}_{\text{base}} \leftarrow \boldsymbol{\mathcal{R}}_{\text{base}} + \boldsymbol{\mathcal{R}}_{\delta}$.

\begin{theorem}
	\label{thm:2}
	Assume the loss function \(\mathcal{L}\) is \(L\)-smooth. Let \(S_{\text{Joint}}\) and \(S_{\text{Decomp}}\) be the stability losses after one training step with learning rate \(\eta\) under the joint and decomposition strategies, respectively. If \(\boldsymbol{\mathcal{R}}_{\delta}\) is initialized such that \(\|\boldsymbol{\mathcal{R}}_{\delta}\|_2 \ll \|\boldsymbol{\mathcal{R}}_{\text{base}}\|_2\), the decomposition strategy achieves a significantly tighter upper bound on the stability loss. 
\end{theorem}

Beyond the magnitude of updates, the variability of these updates is also crucial for stability.As $\boldsymbol{\mathcal{R}}_{\text{base}}$ specializes across tasks, its elements develop high variance (Figure~\ref{fig:2}), whereas $\boldsymbol{\mathcal{R}}_{\delta}$ begins each task with low variance. To analyze the impact of this difference on the gradients, the following theorem assumes for simplicity that the elements of the residual matrices are independent and identically distributed (i.i.d.) with zero mean. 

\begin{theorem}[Reduced Gradient Variance via Decomposition]
	\label{thm:3}
	Let the elements of $\boldsymbol{\mathcal{R}}_{\text{base}}$ and $\boldsymbol{\mathcal{R}}_{\delta}$ be modeled as independent and identically distributed (i.i.d.) random variables with zero mean and variances $\sigma_{\text{base}}^2$ and $\sigma_{\delta}^2$, respectively. Let $\nabla_{\mathbf{A}}(\boldsymbol{\mathcal{R}})$ denote the gradient w.r.t $\mathbf{A}$ modulated by $\boldsymbol{\mathcal{R}}$. If $\sigma_{\text{base}}^2 > \sigma_{\delta}^2$, then the total variance of the gradient updates modulated by $\boldsymbol{\mathcal{R}}_{\delta}$ is strictly smaller than if they were modulated by $\boldsymbol{\mathcal{R}}_{\text{base}}$. Specifically,
	\begin{equation}
		\mathbb{E}[\|\nabla_{\mathbf{A}}(\boldsymbol{\mathcal{R}}_{\delta})\|_F^2] < \mathbb{E}[\|\nabla_{\mathbf{A}}(\boldsymbol{\mathcal{R}}_{\text{base}})\|_F^2],
	\end{equation}
	and similarly for $\nabla_{\mathbf{B}}$.
\end{theorem}

Theorems~\ref{thm:2} and~\ref{thm:3} show that routing through $\boldsymbol{\mathcal{R}}_{\delta}$ produces updates that are both smaller in magnitude and lower in variance—two complementary properties for stability.

\textbf{Fine-grained subspace gating.}
The decomposition mitigates gradient magnitude but treats all $r^2$ subspaces uniformly. Subspaces with large $|\mathcal{R}_{\text{base},ij}|$ are already heavily used and warrant additional protection. We introduce an importance-aware gate $\boldsymbol{\Gamma} \in [0,1]^{r \times r}$ that suppresses gradient flow into $\boldsymbol{\mathcal{R}}_{\delta}$ at dimensions important to prior tasks, while allowing free updates elsewhere:

\begin{equation}
	\boldsymbol{\mathcal{R}}'_\delta = \mathbf{\Gamma} \odot \boldsymbol{\mathcal{R}}_{\delta} + (\mathbf{1} - \mathbf{\Gamma}) \odot \phi(\boldsymbol{\mathcal{R}}_{\delta}).
\end{equation}
In the forward pass, $\boldsymbol{\mathcal{R}}'_\delta = \boldsymbol{\mathcal{R}}_\delta$ numerically. However, the stop-gradient operation $\phi(\cdot)$ ensures that during backward propagation, the effective gradient flowing to the parameters of $\boldsymbol{\mathcal{R}}_{\delta}$ is scaled by $\mathbf{\Gamma}$. When $\Gamma_{ij} \approx 0$ (important subspace), the gradient is suppressed, thereby protecting the consolidated knowledge by preventing adaptation in that dimension. When $\Gamma_{ij} \approx 1$ (low importance), the gradient flows freely.


\subsection{Implementation: C-LoRA in ViT}

Building on our C-LoRA approach, this section details its application to Vision Transformers (ViT)~\cite{dosovitskiy2020image} for efficient continual learning. 
\begin{figure*}[!t]
	\centering
	\begin{minipage}{\textwidth}
		\centering
		\includegraphics[width=\textwidth, trim=0cm 4cm 0cm 3cm, clip]{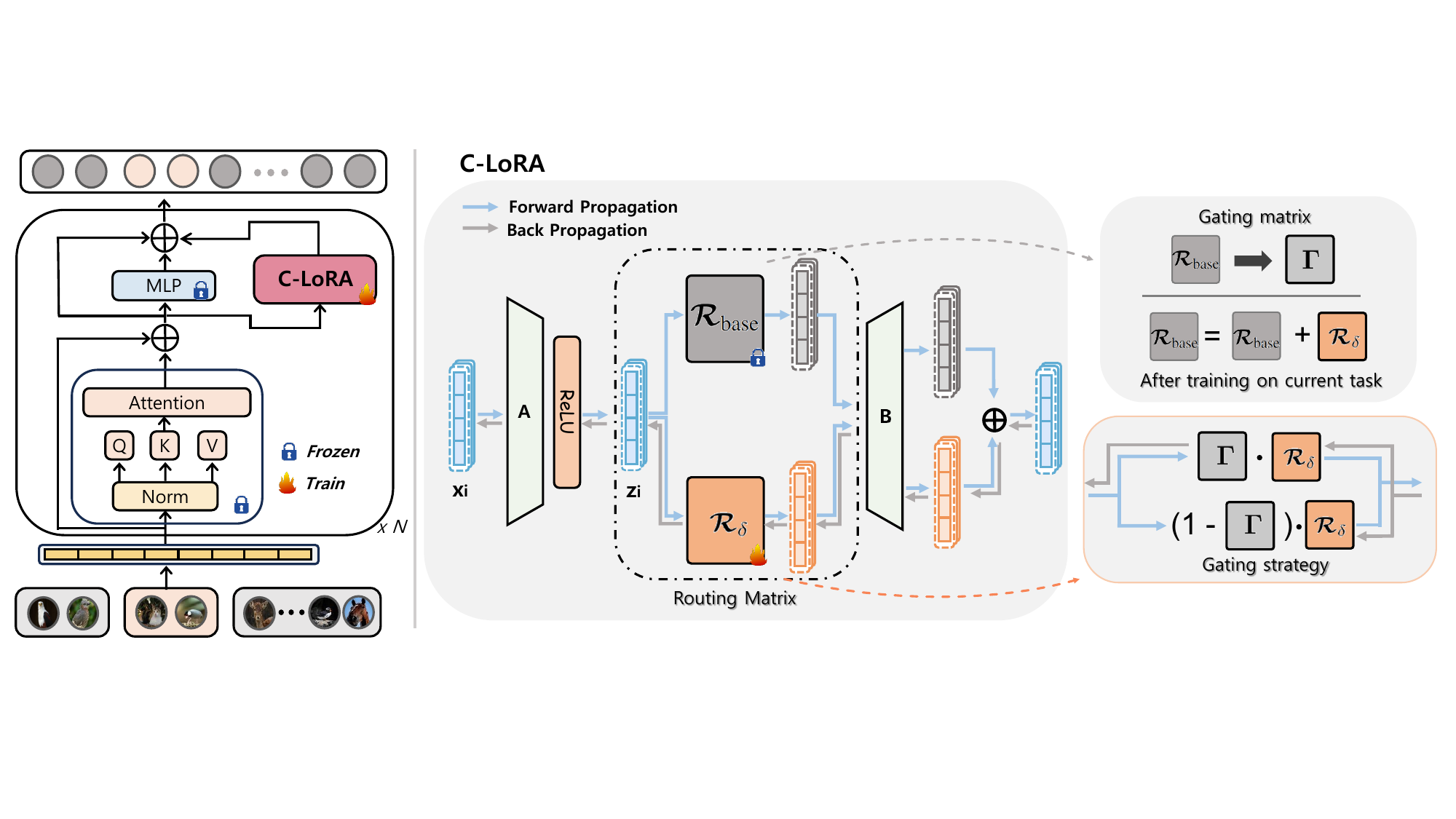}
	\end{minipage}%
	\caption{ Illustration of the Proposed Model. \textbf{Left:} Vision Transformer (ViT) integrated with the C-LoRA module, where the adapter and local classifier are incrementally trained in each session. \textbf{Right:} Proposed architecture mitigates catastrophic forgetting by decoupling \(\boldsymbol{\mathcal{R}}\). This isolates updates within a task-specific component, $\boldsymbol{\mathcal{R}}_\delta$, which in turn drives distributed updates to discover new routing configurations while preserving previously learned knowledge.}
	\label{fig:4} 
\end{figure*}
In a standard MLP block, the input features $\mathbf{x}_i \in \mathbb{R}^d$ are transformed as:
\begin{equation}
	\text{MLP}(\mathbf{x}_i) = \sigma(\mathbf{x}_i \mathbf{W}_1 + \mathbf{b}_1) \mathbf{W}_2 + \mathbf{b}_2,
\end{equation}
where $\mathbf{W}_1 \in \mathbb{R}^{d \times h}$ and $\mathbf{W}_2 \in \mathbb{R}^{h \times d}$ are the weight matrices, $\mathbf{b}_1$ and $\mathbf{b}_2$ are biases, and $\sigma(\cdot)$ is a non-linear activation function, such as ReLU. Directly updating these matrices for new tasks leads to significant parameter overhead and risks overwriting previously learned knowledge.

C-LoRA addresses these challenges by replacing full-rank updates with a low-rank adaptation mechanism. The modified MLP block operates by first projecting the input $\mathbf{x}_i$ into a low-rank subspace using a down-sampling matrix $\mathbf{A} \in \mathbb{R}^{d \times r}$, where $r \ll d$. The intermediate representation is computed as:
\begin{equation}
	\boldsymbol{z}_i = \mathbf{x}_i \mathbf{A},
\end{equation}
where $\mathbf{A}$ reduces the dimensionality of the input while preserving its essential features.

\textbf{Gating implementation. }
We implement the importance-aware gating matrix $\mathbf{\Gamma}$ to protect critical subspaces identified in $\boldsymbol{\mathcal{R}}_{\text{base}}$. The importance of each subspace is estimated by the magnitude of the corresponding element, $\textbf{I}_{ij} = |{\mathcal{R}}_{\text{base}, ij}|$. After the first task, importance scores are Z-score normalized:
\begin{equation}
	Z_{ij} = \frac{\textbf{I}_{ij} - \mu_I}{\sigma_I},
\end{equation}
where $\mu_I$ and $\sigma_I$ are the mean and standard deviation of the importance matrix $\mathbf{I}$, respectively.

The gating matrix $\mathbf{\Gamma}$ is then computed using an element-wise sigmoid function:
\begin{equation}
	\mathbf{\Gamma} = \text{Sigmoid}(-\beta \cdot \mathbf{Z}).
\end{equation}
Here, $\beta > 0$ controls the sharpness of the gate. For the first task, $\mathbf{\Gamma}$ is initialized as a matrix of ones.

We employ the decoupled gating strategy in the update process. The effective routing component for the current task, $\boldsymbol{\mathcal{R}}'_\delta$, is calculated as:
\begin{equation}
	\boldsymbol{\mathcal{R}}'_\delta = \mathbf{\Gamma} \odot \boldsymbol{\mathcal{R}}_{\delta} + (\mathbf{1} - \mathbf{\Gamma}) \odot \phi(\boldsymbol{\mathcal{R}}_{\delta}),
\end{equation}
where $\phi(\cdot)$ denotes the stop-gradient operator.

The representation $\boldsymbol{z}_i$ is modulated before being up-sampled by $\mathbf{B} \in \mathbb{R}^{r \times d}$. The combined C-LoRA output is expressed as:
\begin{equation}
	\label{eq:9_gated}
	\text{C-LoRA}(\boldsymbol{x}_i)
	\;=\;
	\underbrace{\phi\bigl(\boldsymbol{z}_i\,\boldsymbol{\mathcal{R}}_{\text{base}}\,\mathbf{B}\bigr)}_{\text{stop-gradient for base}}
	\;+\;
	\underbrace{\bigl(\boldsymbol{z}_i\,\boldsymbol{\mathcal{R}}'_\delta\,\mathbf{B}\bigr)}_{\text{gated update}}.
\end{equation}
This mechanism ensures that shared knowledge is preserved while allowing flexibility for task-specific updates. After training on the task is complete, the learned knowledge is consolidated: $\boldsymbol{\mathcal{R}}_{\text{base}} \leftarrow \boldsymbol{\mathcal{R}}_{\text{base}} + \boldsymbol{\mathcal{R}}'_\delta$.

\textbf{Integration into ViT: }
As shown in Figure~\ref{fig:4}, C-LoRA modifies the MLP block in ViT by combining its residual connection, original MLP computation, and the C-LoRA updates. The final output is computed as:
\begin{equation}
	\boldsymbol{out} 
	\;=\;
	\boldsymbol{x}_i + \text{MLP}(\boldsymbol{x}_i)
	\;+\;
	\text{C-LoRA}(\boldsymbol{x}_i).
\end{equation}
This formulation maintains the original representational power of the MLP block while incorporating task-specific adaptability through C-LoRA. Over successive tasks, the accumulated updates in $\boldsymbol{\mathcal{R}}_{\delta }$ enable the model to adapt to new tasks without compromising performance on previously learned ones.

\textbf{Loss function.} Each task is trained with a cosine-similarity cross-entropy loss:
\begin{equation}
	\mathcal{L} = -\frac{1}{N_t} \sum_{j=1}^{N_t} \log \frac{\exp(z_{j,c^*})}{\sum_{\ell \in \mathcal{Y}(t)} \exp(z_{j,\ell})},
\end{equation}
where \(N_t\) is the total number of samples in the current task \(t\)'s dataset, and \(\mathcal{Y}(t)\) represents the set of class labels for the current task. The term 
$
z_{j,\ell} = \frac{s (\mathbf{w}_\ell^\top \mathbf{f}_j ) }{\|\mathbf{w}_\ell\| \|\mathbf{f}_j\|}
$ represents the scaled cosine similarity between the feature vector \(\mathbf{f}_j\) of sample \(j\) and the weight vector \(\mathbf{w}_\ell\) of class \(\ell\), with \(s\) being a scaling factor to stabilize optimization.

In class-incremental learning, we typically cannot access raw data from previous tasks. To overcome this, our method re-samples synthetic features, not raw images, from a frozen feature extractor after the initial training is complete. We do this by using the mean and covariance of each class's features~\cite{zhang2023slca} to generate representative feature vectors. This allows us to align the classification layer to the new class distribution while keeping the backbone frozen. This approach ensures that the model retains its discriminative power on old tasks without requiring direct access to their data.

\section{Experiments}

\subsection{Experimental Setup}

\textbf{Datasets.} Following previous studies~\cite{mcdonnell2024ranpac,zhang2023slca,tan2024semantically}, we evaluate our method on four primary benchmark datasets. 
CIFAR-100~\cite{krizhevsky2009learning} consists of 100 classes with 60,000 images, approximately 50,000 for training (83\%) and 10,000 for testing (17\%). 
ImageNet-A~\cite{hendrycks2021natural} includes 200 classes and roughly 7,500 test samples. 
CUB-200~\cite{wah2011caltech} contains 11,788 bird images across 200 subclasses, with about 6,000 for training (51\%) and 5,800 for testing (49\%). 
Car196~\cite{krause20133d} comprises 196 car models with 16,185 images, split into around 8,150 for training (50\%) and 8,035 for testing (50\%). 
For the continual learning setting, these datasets are divided into a sequence of tasks (incremental sessions). We evaluate performance under three scenarios: 5, 10, and 20 incremental sessions.

\noindent\textbf{Implementation Details.} Consistent with recent studies~\cite{mcdonnell2024ranpac,zhang2023slca}, we use the ViT-B/16 model pre-trained on ImageNet-21K~\cite{ridnik2021imagenet} as the frozen backbone. During each incremental session, we train the C-LoRA module and the classifier using the SGD optimizer with a learning rate of 0.01, coupled with a cosine annealing scheduler starting at an initial value of 0.0005. The batch size is set to 48. The plasticity component of the routing matrix, $\boldsymbol{\mathcal{R}}_{\delta}$, is initialized near zero at the beginning of each new session. Baseline methods using LoRA adapters \cite{zhou2024expandable,liang2024inflora} used their recommended ranks, and C-LoRA adopts a fixed rank of $r = 20$ throughout all experiments.

\noindent\textbf{Evaluation Metrics.} We use two standard metrics to assess continual learning performance:
\begin{itemize}
	\item \textbf{Last-Acc (\%):} The average classification accuracy across all classes seen so far, measured after the final incremental session.
	\item \textbf{Inc-Acc (\%):} The average of the Last-Acc values measured at the end of each incremental session. This metric provides a holistic view of the model's performance throughout the entire learning process.
\end{itemize}
All experiments are repeated with three different random seeds, and we report the mean and standard deviation of the results. "Joint-Training" represents the upper bound performance.

\subsection{Comparison with State-of-the-Art Methods}

\begin{figure*}[h]
	\centering
	\begin{minipage}{0.95\textwidth}
		\centering
		\includegraphics[width=\textwidth, trim=0.5cm 10.5cm 1cm 1.8cm, clip]{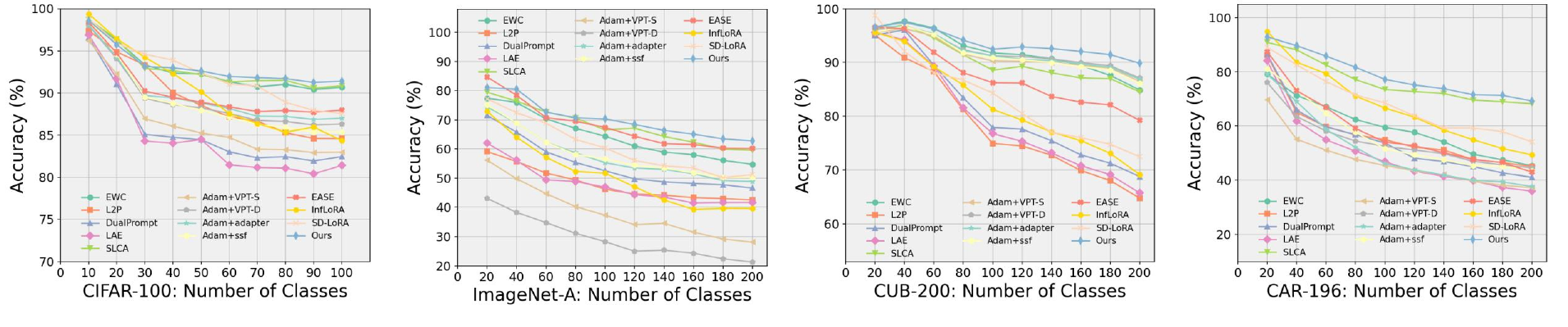}
	\end{minipage}%
	\caption{Accuracy performance during the 10 incremental sessions on CIFAR-100, ImageNet-A, CUB-200, and CAR196.}
	\label{fig:6} 
\end{figure*}

Table~\ref{tab:2} and Figure~\ref{fig:6} present the primary comparison under the 10 session setting, which balances task granularity and sequence length. C-LoRA shows competitive results on all four datasets. The advantage is particularly pronounced on fine-grained benchmarks: on CUB-200, C-LoRA achieves 89.93\% Last-Acc, substantially exceeding Adam+adapter (87.04\%) and SLCA (84.55\%), demonstrating that our stability-plasticity mechanism scales effectively to longer task sequences. Full results under the 5 and 20 session settings are provided in Appendix.

\begin{table*}[h]
	\centering
	\caption{Accuracy comparison of different continual learning methods across four benchmark datasets, divided into 10 incremental sessions. }
	\resizebox{0.95\textwidth}{!}{%
		\begin{tabular}{lcccccccc}
			\toprule
			\textbf{Method} & \multicolumn{2}{c}{\textbf{Split CIFAR-100}} & \multicolumn{2}{c}{\textbf{Split ImageNet-A}} & \multicolumn{2}{c}{\textbf{Split CUB-200}} & \multicolumn{2}{c}{\textbf{Split CAR196}} \\
			\cmidrule(lr){2-3} \cmidrule(lr){4-5} \cmidrule(lr){6-7} \cmidrule(lr){8-9}
			& Last-Acc (\%) & Inc-Acc (\%) & Last-Acc (\%) & Inc-Acc (\%) & Last-Acc (\%) & Inc-Acc (\%) & Last-Acc (\%) & Inc-Acc (\%) \\
			\midrule
			Joint-Training & 93.22 & - & 79.61 & - & 90.78 & - & 78.66 & - \\
			\midrule
			EWC~\cite{kirkpatrick2017overcoming} & 90.59 $\pm$ 0.63 & 93.80 $\pm$ 0.97 & 56.24 $\pm$ 1.56 & 66.30 $\pm$ 1.71 & 85.10 $\pm$ 0.31 & \underline{92.36 $\pm$ 0.51} & 45.58 $\pm$ 0.19 & 60.41 $\pm$ 0.94 \\
			L2P~\cite{wang2022learning} & 84.17 $\pm$ 0.91 & 88.20 $\pm$ 1.30 & 43.47 $\pm$ 0.89 & 50.52 $\pm$ 2.29 & 67.13 $\pm$ 2.03 & 79.62 $\pm$ 1.56 & 45.81 $\pm$ 0.98 & 57.85 $\pm$ 1.92 \\
			DualPrompt~\cite{wang2022dualprompt} & 81.70 $\pm$ 1.07 & 86.42 $\pm$ 0.98 & 46.41 $\pm$ 0.20 & 55.95 $\pm$ 1.34 & 68.68 $\pm$ 0.35 & 80.75 $\pm$ 0.87 & 39.36 $\pm$ 2.32 & 53.93 $\pm$ 2.40 \\
			SLCA~\cite{zhang2023slca} & \underline{91.07 $\pm$ 0.45} & \underline{94.06 $\pm$ 1.07} & 58.96 $\pm$ 0.88 & 66.39 $\pm$ 2.16 & 84.55 $\pm$ 0.19 & 90.69 $\pm$ 0.65 & \underline{66.49 $\pm$ 2.87} & \underline{76.43 $\pm$ 0.61} \\
			Adam+VPT-S~\cite{zhou2023revisiting} & 68.67 $\pm$ 12.40 & 75.22 $\pm$ 9.72 & 39.08 $\pm$ 9.51 & 49.08 $\pm$ 9.11 & 80.03 $\pm$ 11.42 & 86.60 $\pm$ 9.14 & 31.63 $\pm$ 10.15 & 41.78 $\pm$ 10.92 \\
			Adam+VPT-D~\cite{zhou2023revisiting} & 84.61 $\pm$ 3.01 & 88.83 $\pm$ 2.70 & 37.00 $\pm$ 13.81 & 46.29 $\pm$ 14.94 & 86.73 $\pm$ 0.33 & 91.91 $\pm$ 0.15 & 38.04 $\pm$ 12.80 & 48.00 $\pm$ 14.05 \\
			Adam+adapter~\cite{zhou2023revisiting} & 87.32 $\pm$ 0.28 & 91.21 $\pm$ 1.35 & 48.94 $\pm$ 0.10 & 58.95 $\pm$ 1.38 & \underline{87.04 $\pm$ 0.13} & 92.13 $\pm$ 0.27 & 37.65 $\pm$ 0.09 & 49.22 $\pm$ 1.66 \\
			Adam+ssf~\cite{zhou2023revisiting} & 85.26 $\pm$ 0.40 & 89.92 $\pm$ 0.84 & 49.60 $\pm$ 0.80 & 58.90 $\pm$ 2.22 & 86.32 $\pm$ 0.43 & 91.80 $\pm$ 0.55 & 42.83 $\pm$ 0.72 & 53.97 $\pm$ 1.94 \\
			EASE ~\cite{zhou2024expandable} & 87.53 $\pm$ 0.38 & 91.67 $\pm$ 1.04 & \underline{60.74 $\pm$ 0.93} & \underline{69.43 $\pm$ 1.31} & 81.55 $\pm$ 2.38 & 88.63 $\pm$ 1.84 & 44.07 $\pm$ 0.95 & 57.88 $\pm$ 1.12 \\
			InfLoRA~\cite{liang2024inflora} & 84.31 $\pm$ 0.28 & 90.31 $\pm$ 0.28 & 40.42 $\pm$ 1.15 & 52.20 $\pm$ 1.51 & 69.58 $\pm$ 0.78 & 81.57 $\pm$ 0.43 & 51.45 $\pm$ 3.36 & 68.05 $\pm$ 0.62 \\
			SD-LoRA~\cite{wu2025sd} & 86.65 $\pm$ 1.54 & 90.78 $\pm$ 1.25 & 44.11 $\pm$ 9.81 & 53.93 $\pm$ 9.59 & 71.33 $\pm$ 1.21 & 82.69 $\pm$ 0.53 & 52.47 $\pm$ 1.72 & 65.16 $\pm$ 2.72 \\
			\midrule
			\textbf{C-LoRA (Ours)} & \textbf{91.70 $\pm$ 0.14} & \textbf{94.24 $\pm$ 0.85} & \textbf{63.86 $\pm$ 0.84} & \textbf{71.18 $\pm$ 2.10} & \textbf{89.93 $\pm$ 0.49} & \textbf{93.48 $\pm$ 0.11} & \textbf{70.73 $\pm$ 1.68} & \textbf{78.86 $\pm$ 1.32} \\
			\bottomrule
		\end{tabular}%
	}
	\label{tab:2}
\end{table*}

A key advantage of C-LoRA is its \emph{fixed} parameter budget. As shown in Table~\ref{tab:param_comparison}, LoRA-expansion methods incur parameter costs that grow linearly with the number of tasks (e.g., EASE requires 23.6M total expanded parameters over 20 tasks), whereas C-LoRA maintains a constant overhead parameter regardless of sequence length. This fixed budget also eliminates the need for module selection or fusion at inference time, thereby simplifying deployment.

\begin{table}[h]
	\centering
	\caption{Comparison of parameter efficiency for different LoRA-based continual learning methods under a 20-task setting. C-LoRA maintains a fixed parameter budget, while others grow linearly with the number of tasks.}
	\label{tab:param_comparison}
	\small 
	\begin{tabular}{lcc}
		\toprule
		\textbf{Method} & \textbf{Total Expanded Parameters (M)} & \textbf{Trainable Parameters per Task (M)} \\
		\midrule
		EASE~& 23.6 & 1.18 \\
		InfLoRA~& 7.4 & 0.37 \\
		SD-LoRA & 7.4 & 0.37 \\
		\midrule
		\textbf{C-LoRA (Ours)} & \textbf{0.36} & \textbf{0.36} \\
		\bottomrule
	\end{tabular}
\end{table}

\subsection{Ablation Studies}

\begin{figure}[!h]
	\centering
	\begin{minipage}{0.85\textwidth}
	\includegraphics[width=\columnwidth]{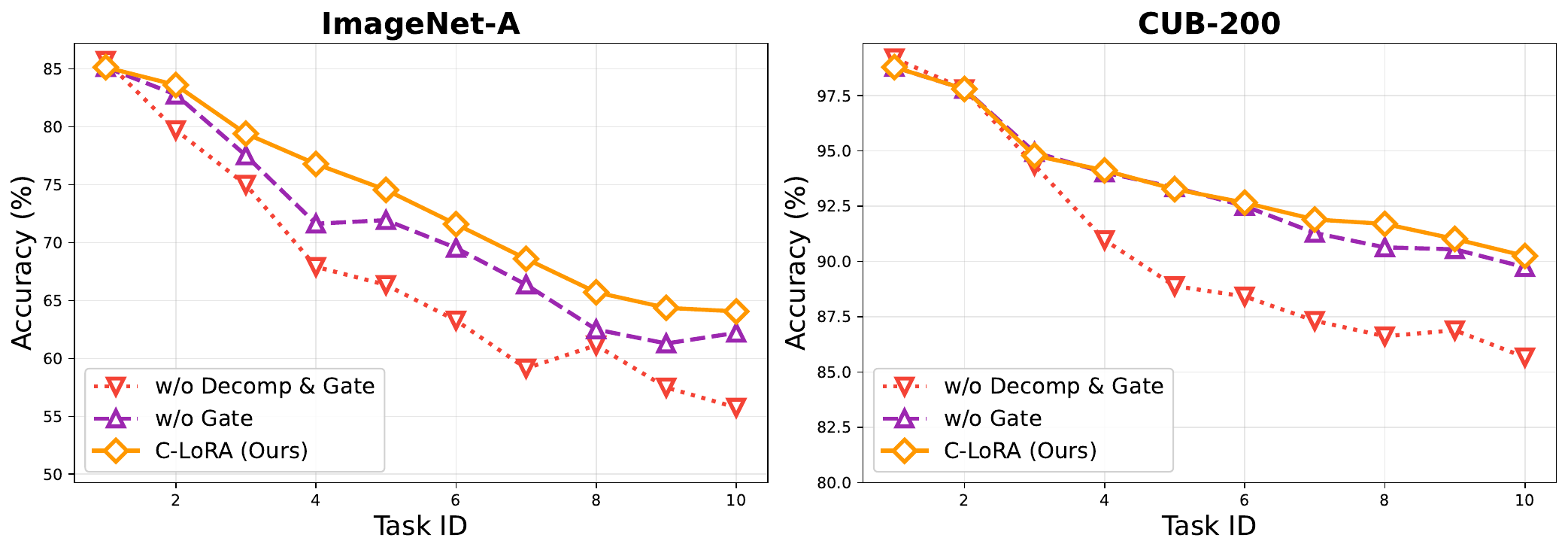}
	\centering
	\end{minipage}
	\caption{Per-task accuracy on ImageNet-A (left) and CUB-200 (right).}
	\label{fig:ablation_acc}
\end{figure}

To ensure controlled comparisons across ablation variants, all ablation experiments are conducted with a single fixed random seed. 
Table~\ref{tab:ablation_main} ablates the two core components of C-LoRA on ImageNet-A and CUB-200 (10 sessions): the decomposition of $\boldsymbol{\mathcal{R}}$ into $\boldsymbol{\mathcal{R}}_{\text{base}}$ and $\boldsymbol{\mathcal{R}}_{\delta}$, and the importance-aware gating $\mathbf{\Gamma}$. We additionally report Backward Transfer (BWT), where higher values indicate less forgetting. Removing both components causes severe forgetting (BWT: $-28.90$ on ImageNet-A, $-11.78$ on CUB-200), confirming that unconstrained routing updates overwrite prior knowledge. 
The decomposition alone substantially recovers performance, as freezing $\boldsymbol{\mathcal{R}}_{\text{base}}$ shields prior subspaces from gradient interference.
Adding gating further reduces forgetting by providing fine-grained, per-subspace control, and the progressive benefit of each component is shown in Figure~\ref{fig:ablation_acc}.

\begin{table}[!h]
	\centering
	\caption{Ablation study on the core components of C-LoRA (10 sessions).}
	\label{tab:ablation_main}
	\small
	\setlength{\tabcolsep}{4pt}
	\begin{tabular}{lcccccc}
		\toprule
		\multirow{2}{*}{\textbf{Method}}
		& \multicolumn{3}{c}{\textbf{ImageNet-A}}
		& \multicolumn{3}{c}{\textbf{CUB-200}} \\
		\cmidrule(lr){2-4}\cmidrule(lr){5-7}
		& Last-Acc (\%) & Inc-Acc (\%) & BWT
		& Last-Acc (\%) & Inc-Acc (\%) & BWT \\
		\midrule
		w/o Decomp \& Gate
		& 55.76 & 67.15 & $-28.90$
		& 85.67 & 90.61 & $-11.78$ \\
		w/o Gate
		& 62.21 & 71.09 & $-13.33$
		& 89.74 & 93.36 & $-4.47$ \\
		\textbf{C-LoRA (Ours)}
		& \textbf{64.06} & \textbf{73.38} & $\mathbf{-9.77}$
		& \textbf{90.25} & \textbf{93.63} & $\mathbf{-3.78}$ \\
		\bottomrule
	\end{tabular}
\end{table}

To isolate the contribution of the C-LoRA module from that of the feature resampling strategy (Stage 2), Table~\ref{tab:lora_ablation} compares C-LoRA against EASE, InfLoRA, and SD-LoRA under the 10 session setting on CIFAR-100 and ImageNet-R~\cite{boschini2022transfer}.
The variant without Stage 2 ("Ours w/o stage 2") already surpasses all LoRA-expansion baselines on both datasets, achieving 88.49\% on CIFAR-100 compared to 87.53\% for EASE. 
This demonstrates that the performance gains of C-LoRA stem primarily from the design of the routing matrix rather than from bias correction alone.
Incorporating Stage 2 yields additional improvements (91.70\% on CIFAR-100; 77.94\% on ImageNet-R), confirming that feature resampling further reduces the recency bias introduced by class-incremental training.

\begin{table}[!h]
	\centering
	\caption{Comparison with LoRA-expansion methods and ablation on the feature resampling strategy (Stage 2) on CIFAR-100 and ImageNet-R (10 sessions).}
	\begin{tabular}{lcccc}
		\toprule
		\textbf{Method} & \multicolumn{2}{c}{\textbf{Split CIFAR-100}} & \multicolumn{2}{c}{\textbf{Split ImageNet-R}} \\
		\cmidrule(lr){2-3} \cmidrule(lr){4-5}
		& Last-Acc (\%) & Inc-Acc (\%) & Last-Acc (\%) & Inc-Acc (\%) \\
		\midrule
		\textit{LoRA-Expansion Methods} \\
		EASE & 87.53 $\pm$ 0.38 & 91.67 $\pm$ 1.04 & 74.52 $\pm$ 0.93 & 80.22 $\pm$ 1.31 \\
		InfLoRA & 84.31 $\pm$ 0.28 & 90.31 $\pm$ 0.28 & 74.75 $\pm$ 0.64 & 80.67 $\pm$ 0.55 \\
		SD-LoRA & 86.65 $\pm$ 1.54 & 90.78 $\pm$ 1.25 & {75.72 $\pm$ 1.15} & {80.39 $\pm$ 0.82} \\
		\midrule
		\textit{C-LoRA (Ours)} \\
		Ours w/o stage 2 & 88.49 $\pm$ 0.24 & 92.19 $\pm$ 1.24 & 76.79 $\pm$ 0.21 & 82.22 $\pm$ 0.59 \\
		\textbf{Ours (Full)} & \textbf{91.70  $\pm$ 0.14} & \textbf{94.24 $\pm$ 0.85} & \textbf{77.94} $\pm$ 0.41 & \textbf{82.99} $\pm$ 0.37 \\
		\bottomrule
	\end{tabular}
	
	\label{tab:lora_ablation}
\end{table}

\section{Conclusion}
In this work, we presented Continual Low-Rank Adaptation (C-LoRA), a novel framework that addresses catastrophic forgetting in pre-trained visual models during sequential task adaptation. By employing a learnable routing matrix $\boldsymbol{\mathcal{R}}$, strategically decomposed to manage shared and task-specific knowledge, C-LoRA significantly enhances knowledge retention. Our extensive empirical evaluations on diverse benchmarks demonstrated C-LoRA's competitive performance.

\section{Acknowledgement}
This work is supported by the National Natural Science Foundation of China (No.62432006, 62276159) and the Fundamental Research Program of Shanxi Province (No. 202303021223004).

\bibliographystyle{unsrtnat} 
\bibliography{example_paper}


\appendix
\section{Mathematical Proofs}
\label{app:proofs} 

This appendix provides the detailed proofs for theorems presented in the main text. 

\subsection{Proof of Theorem 1}
\label{app:proof_thm31} 


\noindent \emph{Proof of Theorem 1. } We aim to establish an upper bound on the Frobenius norm of the gradient $\nabla_{\mathbf{A}}\mathcal{L} = \frac{\partial \mathcal{L}}{\partial \mathbf{A}}$. The gradient is given by $\nabla_{\mathbf{A}}\mathcal{L} = \boldsymbol{G} \mathbf{B}^\top \boldsymbol{\mathcal{R}}^\top$. We invoke the property for the Frobenius norm, which states that for conformable matrices $X$ and $Y$, $\|XY\|_F \leq \|X\|_F \|Y\|_2$, where $\|\cdot\|_2$ denotes the spectral norm.
\begin{equation}
	\begin{aligned}
		\|\nabla_{\mathbf{A}}\mathcal{L}\|_F &= \|\boldsymbol{G} (\boldsymbol{\mathcal{R}}\mathbf{B})^\top\|_F \\
		&\leq \|\boldsymbol{G}\|_F \|\boldsymbol{\mathcal{R}}\mathbf{B}\|_2.
	\end{aligned}
\end{equation}
By the assumption that $\|\boldsymbol{G}\|_F \leq G_{\text{max}}$, and utilizing the sub-multiplicativity property of the spectral norm ($\|\boldsymbol{\mathcal{R}}\mathbf{B}\|_2 \leq \|\boldsymbol{\mathcal{R}}\|_2 \|\mathbf{B}\|_2$), we obtain:
\begin{equation}
	\|\nabla_{\mathbf{A}}\mathcal{L}\|_F \leq G_{\text{max}} \|\boldsymbol{\mathcal{R}}\|_2 \|\mathbf{B}\|_2.
\end{equation}
Since the spectral norm is bounded by the Frobenius norm, $\|\mathbf{B}\|_2 \leq \|\mathbf{B}\|_F$. By the hypothesis that $\|\mathbf{B}\|_F \leq M$, we conclude:
\begin{equation}
	\|\nabla_{\mathbf{A}}\mathcal{L}\|_F \leq G_{\text{max}} M \|\boldsymbol{\mathcal{R}}\|_2.
\end{equation}
The bound for $\nabla_{\mathbf{B}}\mathcal{L}$ is derived symmetrically by considering $\|\boldsymbol{\mathcal{R}}^\top \mathbf{A}^\top \boldsymbol{G}\|_F$, utilizing the assumption $\|\mathbf{A}\|_F \leq M$, and the fact that $\|\boldsymbol{\mathcal{R}}^\top\|_2 = \|\boldsymbol{\mathcal{R}}\|_2$.\hfill $\square$\\

\subsection{Proof of Theorem 2}
\noindent \emph{Proof of Theorem 2. }Let $\mathcal{L}_{old}(\theta) = \mathbb{E}_{D_{old}}[\mathcal{L}(\theta)]$. The stability loss is $S(\theta_t, \theta_{t+1}) = \mathcal{L}_{old}(\theta_{t+1}) - \mathcal{L}_{old}(\theta_t)$. Since $\mathcal{L}_{old}$ is $L$-smooth, we can bound the stability loss using the Descent Lemma (derived from the second-order Taylor expansion):
\begin{equation}
	S(\theta_t, \theta_{t+1}) \leq \langle \nabla \mathcal{L}_{old}(\theta_t), \Delta\theta \rangle + \frac{L}{2}\|\Delta\theta\|_F^2,
\end{equation}
where $\Delta\theta = \theta_{t+1}-\theta_t$ is the parameter update resulting from a single gradient descent step on the new task loss $\mathcal{L}_{new}$, i.e., $\Delta\theta = -\eta \nabla_{\theta}\mathcal{L}_{new}$.

We compare the magnitude of the updates to the shared parameters $\mathbf{A}$ and $\mathbf{B}$.

\textit{Case 1: Joint Strategy.} The update is modulated by $\boldsymbol{\mathcal{R}} = \boldsymbol{\mathcal{R}}_{\text{base}} + \boldsymbol{\mathcal{R}}_{\delta}$. Applying Theorem 1:
\begin{equation}
	\|\Delta\mathbf{A}_{Joint}\|_F = \eta \|\nabla_{\mathbf{A}}\mathcal{L}_{new}\|_F \leq \eta G_{max} M \|\boldsymbol{\mathcal{R}}_{\text{base}} + \boldsymbol{\mathcal{R}}_{\delta}\|_2.
\end{equation}
\textit{Case 2: Decomposition Strategy.} The gradient flow through $\boldsymbol{\mathcal{R}}_{\text{base}}$ is stopped, so the update is modulated only by $\boldsymbol{\mathcal{R}}_{\delta}$:
\begin{equation}
	\|\Delta\mathbf{A}_{Decomp}\|_F \leq \eta G_{max} M \|\boldsymbol{\mathcal{R}}_{\delta}\|_2.
\end{equation}
By the hypothesis that $\|\boldsymbol{\mathcal{R}}_{\delta}\|_2 \ll \|\boldsymbol{\mathcal{R}}_{\text{base}}\|_2$, it follows immediately that $\|\Delta\mathbf{A}_{Decomp}\|_F$ is significantly smaller than $\|\Delta\mathbf{A}_{Joint}\|_F$. The same holds for $\Delta\mathbf{B}$.

Consequently, the total parameter shift $\|\Delta\theta_{Decomp}\|_F$ is significantly smaller than $\|\Delta\theta_{Joint}\|_F$. Substituting this into the $L$-smoothness bound, both the first-order term (bounded via Cauchy-Schwarz by $\|\nabla \mathcal{L}_{old}(\theta_t)\|_F \|\Delta\theta\|_F$) and the second-order term ($\frac{L}{2}\|\Delta\theta\|_F^2$) are substantially tighter for the decomposition strategy, establishing a significantly improved upper bound on $S_{Decomp}$.\hfill $\square$\\

\subsection{Proof of Theorem 3}
\noindent \emph{Proof of Theorem 3.} We analyze the variability of the gradients by examining their total variance under the defined stochastic model. Since the entries of $\boldsymbol{\mathcal{R}}$ are assumed to have zero mean, the expected gradient $\mathbb{E}[\nabla_{\mathbf{A}}(\boldsymbol{\mathcal{R}})]$ is the zero matrix. The total variance (the sum of the variances of all elements) is therefore equivalent to the expected squared Frobenius norm, $\mathbb{E}[\|\nabla_{\mathbf{A}}(\boldsymbol{\mathcal{R}})\|_F^2]$.

Let $\boldsymbol{\mathcal{R}} \in \mathbb{R}^{r \times r}$ be a random matrix with i.i.d. entries such that $\mathbb{E}[\mathcal{R}_{ij}] = 0$ and $\mathrm{Var}(\mathcal{R}_{ij}) = \sigma^2$. The gradient is $\nabla_{\mathbf{A}}(\boldsymbol{\mathcal{R}}) = \boldsymbol{G} \mathbf{B}^\top \boldsymbol{\mathcal{R}}^\top$. Let $\mathbf{M} = \boldsymbol{G} \mathbf{B}^\top$. We treat $\mathbf{M}$ as fixed.

The $(i, j)$-th element of the gradient is $[\nabla_{\mathbf{A}}(\boldsymbol{\mathcal{R}})]_{ij} = \sum_{k=1}^{r} M_{ik} \mathcal{R}_{jk}$. We compute its variance. Due to the independence of the entries of $\boldsymbol{\mathcal{R}}$:
\begin{equation}
	\begin{aligned}
		\mathrm{Var}([\nabla_{\mathbf{A}}(\boldsymbol{\mathcal{R}})]_{ij}) &= \mathrm{Var}\left(\sum_{k=1}^{r} M_{ik} \mathcal{R}_{jk}\right) \\
		&= \sum_{k=1}^{r} M_{ik}^2 \mathrm{Var}(\mathcal{R}_{jk}) \\
		&= \sigma^2 \sum_{k=1}^{r} M_{ik}^2.
	\end{aligned}
\end{equation}
The total variance is the sum over all entries $(i,j)$. Let $d$ be the row dimension of $\mathbf{A}$ (and $\mathbf{M}$).
\begin{equation}
	\begin{aligned}
		\mathbb{E}[\|\nabla_{\mathbf{A}}(\boldsymbol{\mathcal{R}})\|_F^2] &= \sum_{i=1}^{d} \sum_{j=1}^{r} \mathrm{Var}([\nabla_{\mathbf{A}}(\boldsymbol{\mathcal{R}})]_{ij}) \\
		&= \sum_{i=1}^{d} \sum_{j=1}^{r} \left( \sigma^2 \sum_{k=1}^{r} M_{ik}^2 \right) \\
		&= \sigma^2 \sum_{j=1}^{r} \left( \sum_{i=1}^{d} \sum_{k=1}^{r} M_{ik}^2 \right) \\
		&= \sigma^2 r \|\mathbf{M}\|_F^2.
	\end{aligned}
\end{equation}
By the hypothesis $\sigma_{\text{base}}^2 > \sigma_{\delta}^2$, and assuming $\|\mathbf{M}\|_F^2 > 0$ (non-zero upstream gradient), we have:
\begin{equation}
	\begin{aligned}
		\mathbb{E}[\|\nabla_{\mathbf{A}}(\boldsymbol{\mathcal{R}}_{\delta})\|_F^2] &= \sigma_{\delta}^2 r \|\mathbf{M}\|_F^2 \\
		&< \sigma_{\text{base}}^2 r \|\mathbf{M}\|_F^2 \\
		&= \mathbb{E}[\|\nabla_{\mathbf{A}}(\boldsymbol{\mathcal{R}}_{\text{base}})\|_F^2].
	\end{aligned}
\end{equation}
A symmetric argument holds for $\nabla_{\mathbf{B}}$ by analyzing the variance of $\boldsymbol{\mathcal{R}}^\top (\mathbf{A}^\top \boldsymbol{G})$.\hfill $\square$\\

\section{Additional Experimental Results}

This appendix provides the full set of experimental results omitted from the main text for brevity,
including benchmark comparisons under 5 and 20 session settings.


\begin{table*}[!t]
	\centering
	\caption{Accuracy comparison of different continual learning methods across four benchmark datasets, divided into 5 sessions.}
	\resizebox{\textwidth}{!}{%
		\begin{tabular}{lcccccccc}
			\toprule
			\textbf{Method} & \multicolumn{2}{c}{\textbf{Split CIFAR-100}} & \multicolumn{2}{c}{\textbf{Split ImageNet-A}} & \multicolumn{2}{c}{\textbf{Split CUB-200}} & \multicolumn{2}{c}{\textbf{Split CAR196}} \\
			\cmidrule(lr){2-3} \cmidrule(lr){4-5} \cmidrule(lr){6-7} \cmidrule(lr){8-9}
			& Last-Acc (\%) & Inc-Acc (\%) & Last-Acc (\%) & Inc-Acc (\%) & Last-Acc (\%) & Inc-Acc (\%) & Last-Acc (\%) & Inc-Acc (\%) \\
			\midrule
			Joint-Training & 93.22 & - & 79.61 & - & 90.78 & - & 78.66 & - \\
			\midrule
			EWC~\cite{kirkpatrick2017overcoming} & 91.21 $\pm$ 0.53 & 93.93 $\pm$ 0.67 & 60.13 $\pm$ 2.28 & 68.74 $\pm$ 0.87 & 84.73 $\pm$ 0.74 & 91.07 $\pm$ 0.38 & 51.64 $\pm$ 1.36 & 61.47 $\pm$ 1.10 \\
			L2P ~\cite{wang2022learning} & 84.60 $\pm$ 1.74 & 89.24 $\pm$ 1.27 & 46.94 $\pm$ 0.71 & 53.28 $\pm$ 1.13 & 73.07 $\pm$ 1.74 & 81.99 $\pm$ 1.34 & 56.03 $\pm$ 2.86 & 66.43 $\pm$ 2.40 \\
			DualPrompt~\cite{wang2022dualprompt} & 82.66 $\pm$ 2.00 & 87.23 $\pm$ 0.88 & 50.54 $\pm$ 0.98 & 58.18 $\pm$ 1.30 & 74.28 $\pm$ 0.26 & 83.60 $\pm$ 0.44 & 48.90 $\pm$ 0.76 & 61.83 $\pm$ 0.44 \\
			SLCA~\cite{zhang2023slca} & \underline{92.19 $\pm$ 0.56} & \textbf{94.56 $\pm$ 0.58} & \underline{63.14 $\pm$ 1.39} & 68.65 $\pm$ 0.72 & 86.54 $\pm$ 1.27 & 91.51 $\pm$ 0.96 & \underline{75.73 $\pm$ 1.27} & \underline{82.01 $\pm$ 0.72} \\
			Adam+VPT-S~\cite{zhou2023revisiting} & 84.10 $\pm$ 0.88 & 87.89 $\pm$ 1.49 & 24.49 $\pm$ 2.35 & 31.16 $\pm$ 1.73 & 86.64 $\pm$ 0.17 & 91.25 $\pm$ 0.14 & 30.86 $\pm$ 11.39 & 39.34 $\pm$ 13.77 \\
			Adam+VPT-D~\cite{zhou2023revisiting} & 85.89 $\pm$ 4.11 & 89.41 $\pm$ 3.03 & 48.76 $\pm$ 4.90 & 56.59 $\pm$ 5.78 & 86.67 $\pm$ 0.13 & 91.23 $\pm$ 0.10 & 23.52 $\pm$ 13.70 & 30.80 $\pm$ 17.23 \\
			Adam+adapter~\cite{zhou2023revisiting} & 88.63 $\pm$ 0.13 & 91.98 $\pm$ 0.85 & 49.79 $\pm$ 0.67 & 58.38 $\pm$ 1.33 & \underline{87.25 $\pm$ 0.03} & \underline{91.66 $\pm$ 0.23} & 40.75 $\pm$ 0.29 & 50.07 $\pm$ 1.81 \\
			Adam+ssf~\cite{zhou2023revisiting} & 86.98 $\pm$ 0.26 & 90.91 $\pm$ 0.98 & 52.36 $\pm$ 1.53 & 61.08 $\pm$ 2.60 & 86.73 $\pm$ 0.05 & 91.18 $\pm$ 0.19 & 46.07 $\pm$ 0.66 & 55.58 $\pm$ 1.90 \\
			EASE~\cite{zhou2024expandable} & 89.33 $\pm$ 0.16 & 92.53 $\pm$ 0.92 & 62.92 $\pm$ 0.83 & \underline{70.19 $\pm$ 1.18} & 83.93 $\pm$ 1.48 & 89.55 $\pm$ 0.84 & 54.98 $\pm$ 0.80 & 67.09 $\pm$ 1.08 \\
			InfLoRA~\cite{liang2024inflora} & 87.38 $\pm$ 0.48 & 92.13 $\pm$ 0.08 & 48.87 $\pm$ 1.36 & 59.54 $\pm$ 0.56 & 77.32 $\pm$ 1.61 & 86.28 $\pm$ 0.79 & 66.27 $\pm$ 1.27 & 78.10 $\pm$ 0.48 \\
			SD-LoRA~\cite{wu2025sd} & 88.52 $\pm$ 0.45 & 92.16 $\pm$ 1.16 & 54.86 $\pm$ 1.12 & 62.06 $\pm$ 2.88 & 77.52 $\pm$ 2.28 & 86.21 $\pm$ 1.27 & 66.70 $\pm$ 1.85 & 74.98 $\pm$ 1.63 \\
			\midrule
			\textbf{C-LoRA (Ours)} & \textbf{92.27 $\pm$ 0.25} & \underline{94.49 $\pm$ 0.64} & \textbf{65.28 $\pm$ 0.66} & \textbf{71.52 $\pm$ 2.01} & \textbf{89.67 $\pm$ 0.60} & \textbf{93.15 $\pm$ 0.44} & \textbf{77.07 $\pm$ 0.67} & \textbf{83.10 $\pm$ 1.32}\\
			\bottomrule
		\end{tabular}%
	}
	\label{tab:1}
\end{table*}
\begin{figure*}[!t]
	\centering
	\begin{minipage}{\textwidth}
		\centering
		\includegraphics[width=\textwidth, trim=0.5cm 10.5cm 1cm 1.8cm, clip]{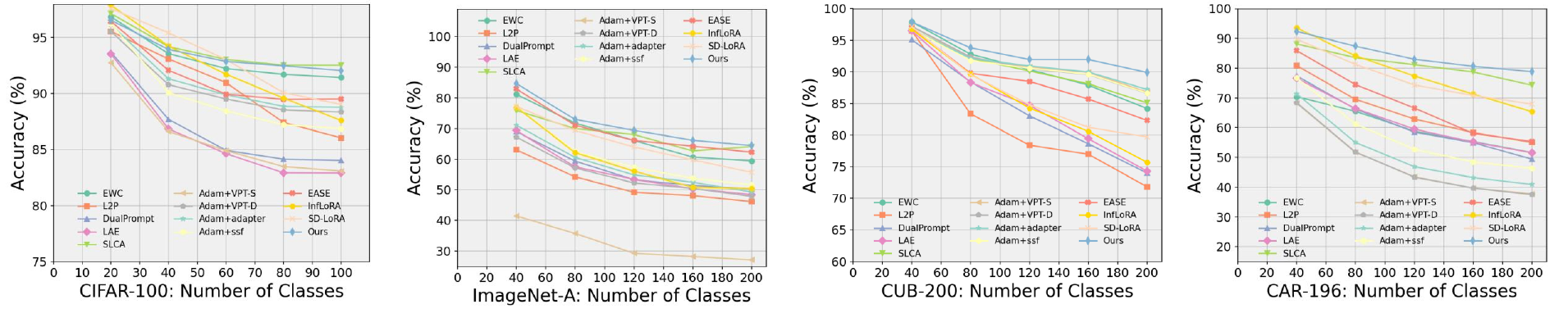}
	\end{minipage}%
	\caption{Accuracy performance during the 5 sessions on CIFAR-100, ImageNet-A, CUB-200, and CAR196.}
	\label{fig:5} 
\end{figure*}

\begin{table*}[!h]
	\centering
	\caption{Accuracy comparison of different continual learning methods across four benchmark datasets, divided into 20 sessions. This setting emphasizes the challenge of longer task sequences.}
	\resizebox{0.95\textwidth}{!}{%
		\begin{tabular}{lcccccccc}
			\toprule
			\textbf{Method} & \multicolumn{2}{c}{\textbf{Split CIFAR-100}} & \multicolumn{2}{c}{\textbf{Split ImageNet-A}} & \multicolumn{2}{c}{\textbf{Split CUB-200}} & \multicolumn{2}{c}{\textbf{Split CAR196}} \\
			\cmidrule(lr){2-3} \cmidrule(lr){4-5} \cmidrule(lr){6-7} \cmidrule(lr){8-9}
			& Last-Acc (\%) & Inc-Acc (\%) & Last-Acc (\%) & Inc-Acc (\%) & Last-Acc (\%) & Inc-Acc (\%) & Last-Acc (\%) & Inc-Acc (\%) \\
			\midrule
			Joint-Training & 93.22 & - & 79.61 & - & 90.78 & - & 78.66 & - \\
			\midrule
			EWC \cite{kirkpatrick2017overcoming} & 89.95 $\pm$ 0.15 & \underline{93.31 $\pm$ 1.02} & 54.62 $\pm$ 1.39 & 63.55 $\pm$ 1.53 & \underline{87.25 $\pm$ 1.00} & \underline{93.45 $\pm$ 0.48} & 44.42 $\pm$ 0.56 & 61.75 $\pm$ 1.06 \\
			L2P~\cite{wang2022learning} & 79.32 $\pm$ 1.49 & 84.10 $\pm$ 1.11 & 40.07 $\pm$ 1.42 & 49.35 $\pm$ 1.37 & 60.30 $\pm$ 3.80 & 73.53 $\pm$ 3.34 & 29.83 $\pm$ 3.31 & 42.82 $\pm$ 1.14 \\
			DualPrompt~\cite{wang2022dualprompt} & 76.57 $\pm$ 0.60 & 82.97 $\pm$ 1.78 & 41.48 $\pm$ 1.63 & 52.84 $\pm$ 1.57 & 65.22 $\pm$ 1.29 & 77.86 $\pm$ 1.31 & 25.11 $\pm$ 2.09 & 40.89 $\pm$ 0.77 \\
			SLCA~\cite{zhang2023slca} & \underline{90.37 $\pm$ 0.56} & \textbf{93.64 $\pm$ 0.99} & 53.76 $\pm$ 5.54 & 61.82 $\pm$ 4.74 & 82.37 $\pm$ 0.42 & 90.12 $\pm$ 0.93 & \underline{56.09 $\pm$ 3.89} & \underline{68.50 $\pm$ 2.42} \\
			Adam+VPT-S~\cite{zhou2023revisiting} & 81.41 $\pm$ 1.00 & 86.48 $\pm$ 1.46 & 40.66 $\pm$ 5.37 & 51.32 $\pm$ 4.68 & 82.42 $\pm$ 5.75 & 89.46 $\pm$ 3.65 & 29.72 $\pm$ 8.31 & 41.08 $\pm$ 8.90 \\
			Adam+VPT-D~\cite{zhou2023revisiting} & 73.60 $\pm$ 15.86 & 79.59 $\pm$ 13.27 & 37.07 $\pm$ 15.26 & 48.41 $\pm$ 15.17 & 86.67 $\pm$ 0.56 & 92.09 $\pm$ 0.87 & 36.76 $\pm$ 6.24 & 49.34 $\pm$ 7.03 \\
			Adam+adapter~\cite{zhou2023revisiting} & 84.86 $\pm$ 0.29 & 89.41 $\pm$ 1.32 & 48.72 $\pm$ 0.07 & 59.53 $\pm$ 1.02 & 86.80 $\pm$ 0.02 & 92.42 $\pm$ 0.31 & 37.69 $\pm$ 0.02 & 50.56 $\pm$ 1.40 \\
			Adam+ssf~\cite{zhou2023revisiting} & 82.96 $\pm$ 1.21 & 87.90 $\pm$ 0.05 & 47.86 $\pm$ 4.13 & 58.58 $\pm$ 4.27 & 86.37 $\pm$ 0.20 & 92.21 $\pm$ 0.63 & 41.98 $\pm$ 0.42 & 54.35 $\pm$ 1.29 \\
			EASE~\cite{zhou2024expandable} & 84.88 $\pm$ 1.09 & 90.08 $\pm$ 1.01 & \underline{55.50 $\pm$ 1.16} & \underline{65.73 $\pm$ 1.77} & 78.26 $\pm$ 1.02 & 87.14 $\pm$ 1.19 & 31.83 $\pm$ 1.38 & 47.90 $\pm$ 1.89 \\
			InfLoRA~\cite{liang2024inflora} & 77.28 $\pm$ 1.19 & 85.82 $\pm$ 0.28 & 30.90 $\pm$ 1.16 & 43.12 $\pm$ 2.98 & 65.24 $\pm$ 2.71 & 77.89 $\pm$ 0.84 & 35.79 $\pm$ 1.56 & 55.10 $\pm$ 2.21 \\
			SD-LoRA~\cite{wu2025sd} & 84.14 $\pm$ 1.39 & 88.73 $\pm$ 2.74 & 40.51 $\pm$ 2.78 & 51.33 $\pm$ 3.18 & 60.64 $\pm$ 1.51 & 74.50 $\pm$ 1.29 & 26.53 $\pm$ 10.48 & 42.02 $\pm$ 13.26 \\
			\midrule
			\textbf{C-LoRA (Ours)} & \textbf{90.70 $\pm$ 0.20} & 93.30 $\pm$ 1.08 & \textbf{59.40 $\pm$ 3.36} & \textbf{69.25 $\pm$ 2.31} & \textbf{90.03 $\pm$ 0.14} & \textbf{93.96 $\pm$ 0.41} & \textbf{62.45 $\pm$ 2.13} & \textbf{73.85 $\pm$ 1.28} \\
			\bottomrule
		\end{tabular}%
	}
	\label{tab:3}
\end{table*}
\begin{figure*}[!h]
	\begin{minipage}{\textwidth}
		\centering
		\includegraphics[width=0.95\textwidth, trim=0.5cm 10.5cm 1cm 1.8cm, clip]{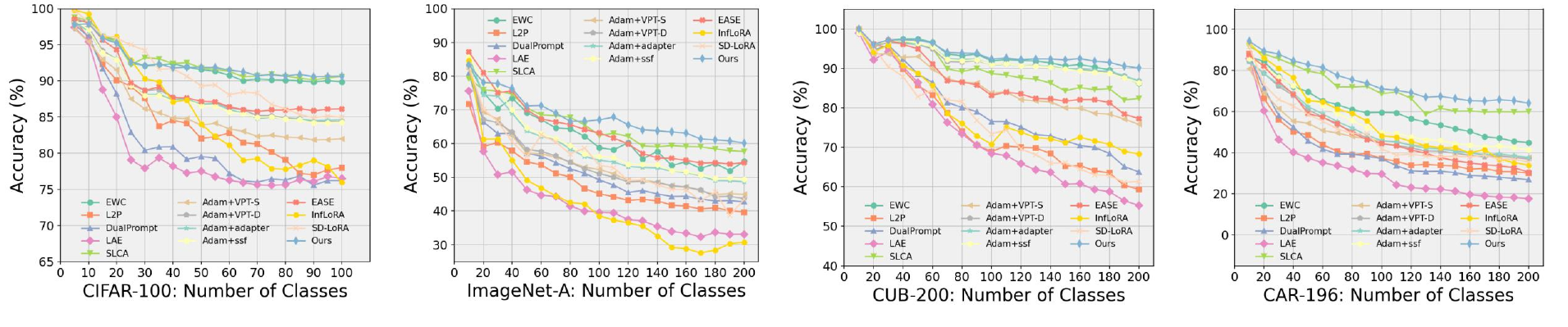}
	\end{minipage}%
	\caption{Accuracy performance during the 20 sessions on CIFAR-100, ImageNet-A, CUB-200, and CAR196.}
	\label{fig:7} 
\end{figure*}

Table~\ref{tab:1} summarises performance when each dataset is divided into 5 sessions.
C-LoRA achieves competitive results across nearly all benchmarks.
On CIFAR‑100, C‑LoRA achieves a Last‑Acc of 92.27\%, slightly ahead of the competitive SLCA (92.19\%). 
On the more challenging ImageNet-A, CUB-200, and CAR196 datasets, C-LoRA consistently outperforms all baselines.
Notably, on CUB-200, C-LoRA (89.67\%) comes within 1.1 percentage points of the Joint-Training upper bound (90.78\%).
The per-session accuracy curves in Figure~\ref{fig:5} further illustrate the stability of C-LoRA throughout the learning process. 
Table~\ref{tab:3} and Figure~\ref{fig:7} further report results under the more demanding 20 session setting. A noteworthy observation is that on CUB-200, C-LoRA's Last-Acc slightly increases with the number of sessions. We attribute this to the substantial overlap between CUB-200's bird categories and the pre-training corpus, reducing the forgetting pressure that would otherwise accumulate over longer task sequences.



\begin{table}[!h]
	\centering
	\caption{CUB-200 per-task accuracy (\%) after learning all tasks.}
	\label{tab:cub_pertask_acc}
	\small
	\setlength{\tabcolsep}{4pt}
	\begin{tabular}{lccccccccccc}
		\toprule
		\textbf{Method}
		& T1 & T2 & T3 & T4 & T5 & T6 & T7 & T8 & T9 & T10 & \textbf{Avg} \\
		\midrule
		w/o Decomp \& Gate
		& 63.2 & 75.4 & 83.9 & 85.8 & 86.2 & 90.3 & 88.4 & 93.2 & 93.7 & 94.8 & 85.5 \\
		w/o Gate
		& 79.8 & 89.4 & 89.0 & 89.5 & 92.2 & 93.4 & 89.3 & 93.6 & 92.0 & 89.6 & 89.8 \\
		\textbf{C-LoRA (Ours)}
		& \textbf{83.8} & \textbf{90.8} & \textbf{90.4} & \textbf{89.1} & \textbf{92.2} & 92.5 & \textbf{90.2} & 92.8 & 91.6 & 89.1 & \textbf{90.3} \\
		\bottomrule
	\end{tabular}
\end{table}

\begin{table}[!h]
	\centering
	\caption{CUB-200 per-task forgetting (\%) after learning all tasks.}
	\label{tab:cub_pertask_forget}
	\small
	\setlength{\tabcolsep}{4pt}
	\begin{tabular}{lccccccccccc}
		\toprule
		\textbf{Method}
		& T1 & T2 & T3 & T4 & T5 & T6 & T7 & T8 & T9 & T10 & \textbf{Avg} \\
		\midrule
		w/o Decomp \& Gate
		& 36.0 & 21.7 & 11.5 & 10.9 & 10.4 & 4.8 & 5.3 & 4.0 & 1.3 & 0.0 & 10.6 \\
		w/o Gate
		& 18.6 & 8.2 & 5.1 & 4.9 & 2.6 & $-0.4$ & 0.5 & $-0.8$ & 1.7 & 0.0 & 4.0 \\
		\textbf{C-LoRA (Ours)}
		& \textbf{14.6} & \textbf{6.8} & \textbf{4.1} & 6.1 & \textbf{2.2} & $\mathbf{-0.9}$ & $\mathbf{-1.3}$ & 1.6 & \textbf{0.8} & 0.0 & \textbf{3.4} \\
		\bottomrule
	\end{tabular}
\end{table}

\begin{figure}[!h]
	\centering
	\includegraphics[width=0.7\columnwidth]{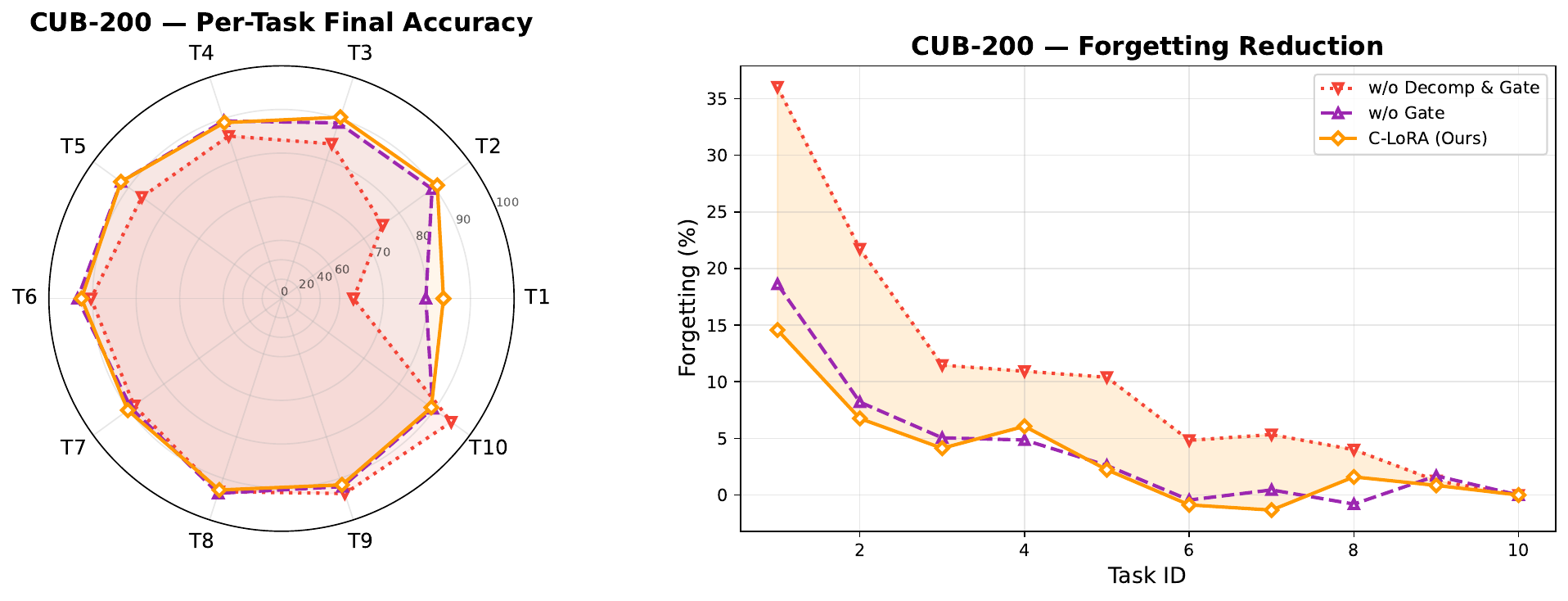}
	\caption{CUB-200: Per-Task Final Accuracy \& Forgetting Reduction}
	\label{fig:cub_forgetting}
\end{figure}

Without decomposition or gating, early tasks suffer the most—T1 loses 36.0\% accuracy as subsequent training freely overwrites shared routing directions.
Routing decomposition halves the average forgetting from 10.6 to 4.0, and C-LoRA reduces it further to 3.4\%. Tables~\ref{tab:cub_pertask_acc} and~\ref{tab:cub_pertask_forget} report per-task accuracy and forgetting on CUB-200 after all 10 sessions are learned.


\label{app:grad_analysis}

To examine the relationship between $\|\boldsymbol{\mathcal{R}}\|_2$ and the gradients on the shared bases described in Theorem~\ref{thm:1}, we log $\|\partial\mathcal{L}/\partial\mathbf{A}\|_F$ and $\|\boldsymbol{\mathcal{R}}\|_2$ at every training step and report the 12-layer mean in Table~\ref{tab:grad_norm} (10 sessions). Without decomposition, $\|\boldsymbol{\mathcal{R}}\|_2$ keeps growing across sessions and the gradient on $\mathbf{A}$ grows with it. With decoupled routing, $\|\boldsymbol{\mathcal{R}}\|_2$ stays close to its initial scale and the gradient on $\mathbf{A}$ is $24\times$ to $8836\times$ smaller.

\begin{table}[!h]
	\centering
	\caption{Gradient norm $\|\partial\mathcal{L}/\partial\mathbf{A}\|_F$ and routing-matrix norm $\|\boldsymbol{\mathcal{R}}\|_2$ across sessions (12-layer mean, 10 sessions). ``w/o Decomp'' denotes the variant without decomposition and gating; Ratio $=$ w/o Decomp / C-LoRA.}
	\label{tab:grad_norm}
	\small
	\setlength{\tabcolsep}{4pt}
	\begin{tabular}{lcccccc}
		\toprule
		\multirow{2}{*}{\textbf{Session}}
		& \multicolumn{3}{c}{$\|\partial\mathcal{L}/\partial\mathbf{A}\|_F$}
		& \multicolumn{3}{c}{$\|\boldsymbol{\mathcal{R}}\|_2$} \\
		\cmidrule(lr){2-4}\cmidrule(lr){5-7}
		& C-LoRA & w/o Decomp & Ratio & C-LoRA & w/o Decomp & Ratio \\
		\midrule
		\multicolumn{7}{l}{\textit{ImageNet-A}} \\
		1 & 0.0013 & 0.0506 & 39$\times$  & 1.09 & 1.56 & 1.43$\times$ \\
		5 & 0.0005 & 0.1118 & 234$\times$ & 1.17 & 2.65 & 2.26$\times$ \\
		9 & 0.0007 & 0.1638 & 236$\times$ & 1.24 & 3.25 & 2.63$\times$ \\
		\midrule
		\multicolumn{7}{l}{\textit{CUB-200}} \\
		1 & $3{\times}10^{-6}$ & 0.0019 & 659$\times$  & 1.03 & 1.51 & 1.47$\times$ \\
		5 & $1{\times}10^{-6}$ & 0.0072 & 4929$\times$ & 1.03 & 2.52 & 2.44$\times$ \\
		9 & $1{\times}10^{-6}$ & 0.0124 & 8836$\times$ & 1.04 & 3.06 & 2.95$\times$ \\
		\midrule
		\multicolumn{7}{l}{\textit{Cars-196}} \\
		1 & 0.0006 & 0.0140 & 24$\times$  & 1.12 & 1.53 & 1.37$\times$ \\
		5 & 0.0005 & 0.0274 & 60$\times$  & 1.31 & 2.49 & 1.90$\times$ \\
		9 & 0.0003 & 0.0373 & 129$\times$ & 1.50 & 3.05 & 2.03$\times$ \\
		\bottomrule
	\end{tabular}
\end{table}

\label{app:controlled}

Table~\ref{tab:controlled} compares C-LoRA with three controlled variants under the same backbone, classifier head, Stage-2 resampling, and protocol as Table~\ref{tab:ablation_main} (10 sessions). \textbf{Vanilla LoRA} uses the standard update $\Delta\mathbf{W}=\mathbf{AB}$ at the same rank. \textbf{w/o Decomp \& Gate} uses a shared $\mathbf{A}\boldsymbol{\mathcal{R}}\mathbf{B}$ with a fully trainable $\boldsymbol{\mathcal{R}}$. \textbf{+ LR scaling} further multiplies the learning rates of $\mathbf{A}$ and $\mathbf{B}$ by $\kappa_t^2$ at each layer, where $\kappa_t=\|\boldsymbol{\mathcal{R}}_{\delta}\|_2/\|\boldsymbol{\mathcal{R}}_{\text{base}}+\boldsymbol{\mathcal{R}}_{\delta}\|_2$, matching the effective step size of decomposition without decomposing $\boldsymbol{\mathcal{R}}$.

\begin{table}[!h]
	\centering
	\caption{Controlled comparison of capacity and learning-rate scaling (10 sessions). ``--'' denotes not evaluated.}
	\label{tab:controlled}
	\small
	\setlength{\tabcolsep}{4pt}
	\begin{tabular}{lcccccc}
		\toprule
		\multirow{2}{*}{\textbf{Method}}
		& \multicolumn{2}{c}{\textbf{ImageNet-A}}
		& \multicolumn{2}{c}{\textbf{CUB-200}}
		& \multicolumn{2}{c}{\textbf{Cars-196}} \\
		\cmidrule(lr){2-3}\cmidrule(lr){4-5}\cmidrule(lr){6-7}
		& Last-Acc & Inc-Acc & Last-Acc & Inc-Acc & Last-Acc & Inc-Acc \\
		\midrule
		Vanilla LoRA
		& 53.85 & -- & 85.33 & -- & 69.38 & -- \\
		w/o Decomp \& Gate
		& 55.76 & 67.15 & 85.67 & 90.61 & 69.69 & 79.19 \\
		\quad + LR scaling
		& 60.11 & 69.85 & 89.99 & -- & 70.44 & 78.98 \\
		\textbf{C-LoRA (Ours)}
		& \textbf{64.06} & \textbf{73.38} & \textbf{90.25} & \textbf{93.63} & \textbf{71.97} & \textbf{80.01} \\
		\bottomrule
	\end{tabular}
\end{table}

Adding a shared routing matrix without decomposition improves over Vanilla LoRA by at most 1.91 points in Last-Acc, so the extra capacity of $\boldsymbol{\mathcal{R}}$ alone does not explain the gains. Learning-rate scaling recovers a large part of the improvement, and C-LoRA still achieves the best accuracy on all three datasets.

\label{app:adalora}

We additionally compare C-LoRA with AdaLoRA~\cite{zhang2023adalora}, obtained by replacing the routing module of C-LoRA with the AdaLoRA parameterization under otherwise identical configurations, and with CL-LoRA~\cite{he2025cl}. Results on ImageNet-A and CUB-200 (10 sessions) are reported in Table~\ref{tab:adalora}.

\begin{table}[!h]
	\centering
	\caption{Comparison with AdaLoRA and CL-LoRA on ImageNet-A and CUB-200 (10 sessions).}
	\label{tab:adalora}
	\small
	\setlength{\tabcolsep}{4pt}
	\begin{tabular}{lcccc}
		\toprule
		\multirow{2}{*}{\textbf{Method}}
		& \multicolumn{2}{c}{\textbf{ImageNet-A}}
		& \multicolumn{2}{c}{\textbf{CUB-200}} \\
		\cmidrule(lr){2-3}\cmidrule(lr){4-5}
		& Last-Acc (\%) & Inc-Acc (\%) & Last-Acc (\%) & Inc-Acc (\%) \\
		\midrule
		AdaLoRA
		& 57.41 & 68.56 & 83.67 & 91.52 \\
		CL-LoRA
		& 58.92 & 70.06 & 80.03 & 87.66 \\
		\textbf{C-LoRA (Ours)}
		& \textbf{64.06} & \textbf{73.38} & \textbf{90.25} & \textbf{93.63} \\
		\bottomrule
	\end{tabular}
\end{table}


\subsection{Computational Resources}
\label{app:compute} 

All experiments were conducted on a GPU server running Ubuntu 20.04.6 LTS. The server was equipped with an Intel Core i9-14900k processor and used a single NVIDIA GeForce RTX 4090 GPU. The implementation was built using the PyTorch framework, version 2.3.1. For the most extensive scenario involving 20 incremental sessions, the average training time required per session on each dataset was approximately as follows: 92 minutes for CIFAR-100, 13 minutes for ImageNet-A, 52 minutes for CUB-200, and 124 minutes for CAR196. These times reflect the average duration to complete the training for one incremental task within the 20 session sequence.

\subsection{Limitations and Future Work}

The theoretical analysis in Theorems~\ref{thm:1}--\ref{thm:3} relies on simplified assumptions, including bounded parameter norms and an i.i.d. zero-mean model for routing matrix elements. These assumptions do not hold exactly during training. Nevertheless, the consistent empirical gains across all benchmarks suggest that the core mechanism—decoupling gradient flow via $\boldsymbol{\mathcal{R}}_{\text{base}}$ and $\boldsymbol{\mathcal{R}}_{\delta}$—is robust beyond the idealized setting. Deriving tighter guarantees under more realistic assumptions is a natural direction for future work.

The experiments in this paper focus on class-incremental image classification with ViT backbones. The routing matrix design is architecture-agnostic in principle, but its effectiveness on other tasks—such as object detection or segmentation—and on other backbone families remains to be verified. Extending C-LoRA to these settings is a direct avenue for future investigation.


\newpage
\section*{NeurIPS Paper Checklist}

\begin{enumerate}
	
	\item {\bf Claims}
	\item[] Question: Do the main claims made in the abstract and introduction accurately reflect the paper's contributions and scope?
	\item[] Answer: \answerYes{} 
	\item[] Justification: The abstract and introduction include the claims made in the paper.
	
	\item {\bf Limitations}
	\item[] Question: Does the paper discuss the limitations of the work performed by the authors?
	\item[] Answer: \answerYes{} 
	\item[] Justification: Limitations concerning theoretical assumptions and empirical scope are discussed in Appendix.
	
	\item {\bf Theory assumptions and proofs}
	\item[] Question: For each theoretical result, does the paper provide the full set of assumptions and a complete (and correct) proof?
	\item[] Answer: \answerYes{} 
	\item[] Justification: Assumptions are stated and acknowledged in Appendix.
	
	\item {\bf Experimental result reproducibility}
	\item[] Question: Does the paper fully disclose all the information needed to reproduce the main experimental results of the paper to the extent that it affects the main claims and/or conclusions of the paper (regardless of whether the code and data are provided or not)?
	\item[] Answer: \answerYes{} 
	\item[] Justification: Methodology, implementation details including hyperparameters and setup, and computational resources are provided.

	\item {\bf Open access to data and code}
	\item[] Question: Does the paper provide open access to the data and code, with sufficient instructions to faithfully reproduce the main experimental results, as described in supplemental material?
	\item[] Answer: \answerYes{} 
	\item[] Justification: We provide the source code in the supplemental material.

	\item {\bf Experimental setting/details}
	\item[] Question: Does the paper specify all the training and test details (e.g., data splits, hyperparameters, how they were chosen, type of optimizer) necessary to understand the results?
	\item[] Answer: \answerYes{} 
	\item[] Justification: Experimental setup, datasets, evaluation protocol, and hyperparameters are detailed in main paper. Compute details are in Appendix.
	
	\item {\bf Experiment statistical significance}
	\item[] Question: Does the paper report error bars suitably and correctly defined or other appropriate information about the statistical significance of the experiments?
	\item[] Answer: \answerYes{} 
	\item[] Justification: Results presented in Tables \ref{tab:3}, \ref{tab:1}, \ref{tab:2} report the mean and standard deviation over 3 independent runs.
	
	\item {\bf Experiments compute resources}
	\item[] Question: For each experiment, does the paper provide sufficient information on the computer resources (type of compute workers, memory, time of execution) needed to reproduce the experiments?
	\item[] Answer: \answerYes{} 
	\item[] Justification: We provide related information in appendix.
	
	\item {\bf Code of ethics}
	\item[] Question: Does the research conducted in the paper conform, in every respect, with the NeurIPS Code of Ethics \url{https://neurips.cc/public/EthicsGuidelines}?
	\item[] Answer: \answerYes{} 
	\item[] Justification: The research focuses on algorithmic development for continual learning using standard public datasets and does not present obvious ethical concerns.

	\item {\bf Broader impacts}
	\item[] Question: Does the paper discuss both potential positive societal impacts and negative societal impacts of the work performed?
	\item[] Answer: \answerYes{} 
	\item[] Justification: Potential positive impacts and broader challenges are discussed at the end of Appendix.
	
	\item {\bf Safeguards}
	\item[] Question: Does the paper describe safeguards that have been put in place for responsible release of data or models that have a high risk for misuse (e.g., pre-trained language models, image generators, or scraped datasets)?
	\item[] Answer: \answerNA{} 
	\item[] Justification: No data or models are released as part of this work, so safeguards for responsible release are not applicable.
	
	\item {\bf Licenses for existing assets}
	\item[] Question: Are the creators or original owners of assets (e.g., code, data, models), used in the paper, properly credited and are the license and terms of use explicitly mentioned and properly respected?
	\item[] Answer: \answerYes{} 
	\item[] Justification: We cite the datasets and baseline papers in the experiment section we used.
	
	\item {\bf New assets}
	\item[] Question: Are new assets introduced in the paper well documented and is the documentation provided alongside the assets?
	\item[] Answer: \answerNA{} 
	\item[] Justification: The paper does not release new assets.
	
	\item {\bf Crowdsourcing and research with human subjects}
	\item[] Question: For crowdsourcing experiments and research with human subjects, does the paper include the full text of instructions given to participants and screenshots, if applicable, as well as details about compensation (if any)? 
	\item[] Answer: \answerNA{} 
	\item[] Justification: The research does not involve crowdsourcing or direct experiments with human subjects.
	
	\item {\bf Institutional review board (IRB) approvals or equivalent for research with human subjects}
	\item[] Question: Does the paper describe potential risks incurred by study participants, whether such risks were disclosed to the subjects, and whether Institutional Review Board (IRB) approvals (or an equivalent approval/review based on the requirements of your country or institution) were obtained?
	\item[] Answer: \answerNA{} 
	\item[] Justification: The research does not involve human subjects requiring IRB approval.
	
	\item {\bf Declaration of LLM usage}
	\item[] Question: Does the paper describe the usage of LLMs if it is an important, original, or non-standard component of the core methods in this research? Note that if the LLM is used only for writing, editing, or formatting purposes and does \emph{not} impact the core methodology, scientific rigor, or originality of the research, declaration is not required.
	\item[] Answer: \answerNA{} 
	\item[] Justification: LLMs were only used for writing, editing, and formatting purposes, and not as a core method in this research.
	
\end{enumerate}

\end{document}